\documentclass[letterpaper]{article} 
\usepackage{aaai25}  
\usepackage{times}  
\usepackage{helvet}  
\usepackage{courier}  
\usepackage[hyphens]{url}  
\usepackage{graphicx} 
\usepackage{natbib}  
\usepackage{caption} 
\usepackage{algorithm}
\usepackage{algorithmic}
\usepackage{multirow}
\usepackage{graphicx}
\usepackage{booktabs}
\usepackage{xcolor}
\usepackage{newfloat}
\usepackage{listings}
\definecolor{green}{HTML}{1b9e77}
\definecolor{red}{HTML}{d95f02}
\DeclareCaptionStyle{ruled}{labelfont=normalfont,labelsep=colon,strut=off} 
\floatstyle{ruled}
\newfloat{listing}{tb}{lst}{}
\floatname{listing}{Listing}
\usepackage{tabularx}
\usepackage{dblfloatfix}
\newcolumntype{Y}{>{\centering\arraybackslash}X}
\newcolumntype{Z}{>{\centering\arraybackslash}X}

\title{DivShift: Exploring Domain-Specific Distribution Shift in Large-Scale, Volunteer-Collected Biodiversity Datasets
}

\author{
    Elena Sierra\textsuperscript{\rm 1,2,3}\equalcontrib, 
    Lauren E. Gillespie\textsuperscript{\rm 1,3,4,5}\equalcontrib, 
    Salim Soltani\textsuperscript{\rm 2}, \\ 
    Moises Exposito-Alonso\textsuperscript{\rm 5,6}, 
    Teja Kattenborn\textsuperscript{\rm 2}
}
\affiliations{
    \textsuperscript{\rm 1}Stanford University 
    Department of Computer Science
    \\
    \textsuperscript{\rm 2}{
    Chair of Sensor-based Geoinformatics (geosense), 
    University of Freiburg
    \\
    } 
    \textsuperscript{\rm 3}Carnegie Science 
    Plant Biology
    \\
    \textsuperscript{\rm 4}Federal University of Minas Gerais 
    Department of Vegetation Biology
    \\
    \textsuperscript{\rm 5}University of California, Berkeley 
    Department of Integrative Biology
    \\
    \textsuperscript{\rm 6}Howard Hughes Medical Institute
    \\
    esierra@cs.stanford.edu, gillespl@cs.stanford.edu
}

\begin{document}

\maketitle

\begin{abstract}
Large-scale, volunteer-collected datasets of community-identified natural world imagery like iNaturalist have enabled marked performance gains for fine-grained visual classification of species using machine learning methods. However, such data---sometimes referred to as citizen science data---are opportunistic and lack a structured sampling strategy. This volunteer-collected biodiversity data contains geographic, temporal, taxonomic, observers, and sociopolitical biases that can have significant effects on biodiversity model performance, but whose impacts are unclear for fine-grained species recognition performance. Here we introduce Diversity Shift (DivShift), a framework for quantifying the effects of domain-specific distribution shifts on machine learning model performance. To diagnose the performance effects of biases specific to volunteer-collected biodiversity data, we also introduce DivShift - North American West Coast (DivShift-NAWC), a curated dataset of almost 7.5 million iNaturalist images across the western coast of North America partitioned across five types of expert-verified bias. We compare species recognition performance across these bias partitions using a diverse variety of species- and ecosystem-focused accuracy metrics. We observe that these biases confound model performance less than expected from the underlying label distribution shift, and that more data leads to better model performance but the magnitude of these improvements are bias-specific. These findings imply that while the structure within natural world images provides generalization improvements for biodiversity monitoring tasks, the biases present in volunteer-collected biodiversity data can also affect model performance; thus these models should be used with caution in downstream biodiversity monitoring tasks.
\end{abstract}

%
\begin{links}
    \link{Code}{github.com/moiexpositoalonsolab/DivShift}
    \link{Dataset}{huggingface.co/datasets/elenagsierra/DivShift-NAWC}
    \link{Extended version}{arxiv.org/abs/2410.19816}
\end{links}

\section{Introduction}
Monitoring biodiversity is vital for understanding the state of the natural world, and frequent and accurate monitoring via automated tools is crucial for guiding decisions to protect the world's ecosystems. Building machine learning tools for this automated monitoring requires large volumes of natural world imagery. In recent years, participatory science applications that enable public volunteers to observe, share, and help identify species in their natural environments have seen a surge in popularity.

These scientific efforts on the part of the general public now mean that large-scale biodiversity image datasets are readily available with the number of observations rapidly approaching the scale of internet-scale image datasets \cite{schuhmann2022laion}. With these finely labeled volunteer-collected datasets, computer vision models have shown impressive improvement in a variety of biodiversity monitoring-related machine learning tasks, including species recognition, species distribution modeling, novel species identification, and visual question answering \cite{iNat2021,crisp, SatBird, sastry2024birdsat, PlantNet, PlantCLEF,deepbiosphere, bioclip}. 
However, the volume of these opportunistic volunteer records comes at a cost: as these observations become easier for the public to collect, sampling becomes unstructured, and injects a variety of biases into these data \cite{arazy2021framework, pernat2021citizen, geldmann2016, isaac2015bias, di2021observing, boakes2010distorted, dimson_who_2023}. These biases mean these volunteer-collected data do not reflect the state of the world's biodiversity in many aspects and present challenges for the general uptake of these unstructured, opportunistic data for biodiversity monitoring \cite{backstrom2024estimating, cooper2014there, callaghan2021large, kishimoto2021covid, johnston2023outstanding, botts2011geographic, deacon2023overcoming, wolf2022citizen}.

\begin{figure} [h]
        \begin{center}
                \includegraphics[width=1\linewidth]{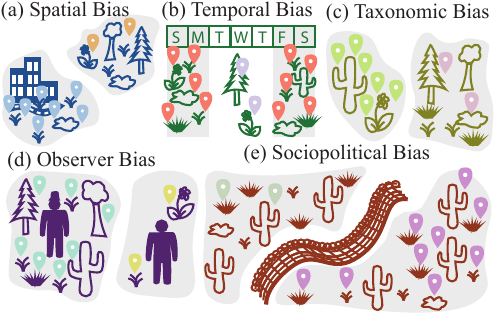}
        \end{center}
\caption{Biases present in biodiversity data include \textbf{(a)} spatial bias, \textbf{(b)} temporal bias, \textbf{(c)} taxonomic bias, \textbf{(d)} observer behavior bias, and \textbf{(e)} sociopolitical bias. 
\label{fig:f1}
}
\end{figure}

To help quantify the effects of these biases on model performance, we introduce Diversity Shift (DivShift), a framework for linking domain shift-driven computer vision model performance disparities to biases manifest in large-scale, volunteer-collected biodiversity datasets. We specifically focus on known biases in these data present across space, time, taxonomy, observers, and sociopolitical boundaries (Fig. ~\ref{fig:f1}). We also introduce a new public biodiversity imagery dataset DivShift-North American West Coast (DivShift-NAWC), a dataset of nearly 7.5 million observations of over 7,500 plant species across the North American West Coast designed to help quantify these disparities in a controlled case study. Performance varies both positively and negatively under these five different domain shifts.
Synthesizing these quantitative results with previous work, we suggest recommendations for downstream use of computer vision models trained on these volunteer-collected biodiversity data.

\section{Related Works}
\subsection{Large-Scale Natural World Imagery Datasets}
Large-scale natural world imagery datasets for training computer vision models for biodiversity monitoring tasks span a variety of modalities, including handheld phone images, high-quality archival and herbaria images, long-distance camera imagery, terrestrial camera traps, ocean sonar cameras, google street view imagery
and remote sensing imagery \cite{iNat2021, crisp, sastry2024birdsat, PlantNet, PlantCLEF, van2018inaturalist, bioclip, de2022herbarium, wang2023bird, BirdCollect, beery2021iwildcam, swanson2015snapshot, cfc2022eccv, AutoArborist, lee2024treedfusionsimulationreadytree, cole2020geolifeclef, deepbiosphere, SatBird, crisp, weinstein2021remote, soltani2024simple}. The app iNaturalist---where users can upload photos of species in their natural environments, identify them, and help identify other observations---has especially seen significant and sustained growth year over year, and now reaches over 40 million observations with nearly 300,000 species observed annually \cite{di2021observing, backstrom2024estimating, dimson_who_2023,inat_review}. 

\subsection{Biases in Volunteer-Collected Biodiversity Datasets}

Collections of large-scale volunteer datasets are subject to social and ecological filters, which inject many types of bias into biodiversity datasets 
\cite{carlen2024framework, isaac2015bias, di2021observing,isaac2014statistics}. In this work, we focus on five kinds of biases common to volunteer-collected biodiversity datasets: spatial, temporal, taxonomic, observer, and sociopolitical (Fig. \ref{fig:f1}). \textbf{Spatial bias} includes observer preferences to sampling easy-to-access green spaces in urban or touristic areas 
\cite{gratzer2021and, mcgoff2017finding, backstrom2024estimating, dimson_who_2023}.  
\textbf{Temporal bias} includes a skew towards more observations on weekends when observers are free from work and during seasons with pleasant weather and attractive appearances of plants \cite{sweet2022covid, sanchez2021differential, crimmins_covid-19_2021, inat_review, courter2013weekend, cooper2014there}. 
\textbf{Taxonomic bias} includes observer preference for identifying larger, exotic, or charismatic species and species that are easy to identify \cite{aristeidou2021exploring, unger2021inaturalist, ward2014understanding, mair2016explaining, mcmullin2022assessment, hochmair2020evaluating, boakes_patterns_2016, deacon2023overcoming, callaghan2021large, stoudt2022identifying}. 
\textbf{Observer bias}
manifests as a small but dedicated group of users that tend to observe more species in more diverse habitats \cite{van2021impact, milanesi2020observer, boakes_patterns_2016, rosenblatt2022highly}. Lastly, \textbf{sociopolitical bias} in who has access to the resources, time, and areas to collect biodiversity observations includes a skew towards whiter and wealthier regions \cite{blake2020demographics, ellis2023historical, mahmoudi2022mapping, chen2022contrasting, burgess2017science, cooper2023equitable,soleri2016finding, pandya2012framework, mac2020citizens,pateman2021diversity}. While these biases are well-documented, their performance effects on fine-grained visual classification of species is not well understood.

\section{The DivShift Framework}
\begin{figure} [t]
        \begin{center}
                \includegraphics[width=1\linewidth]{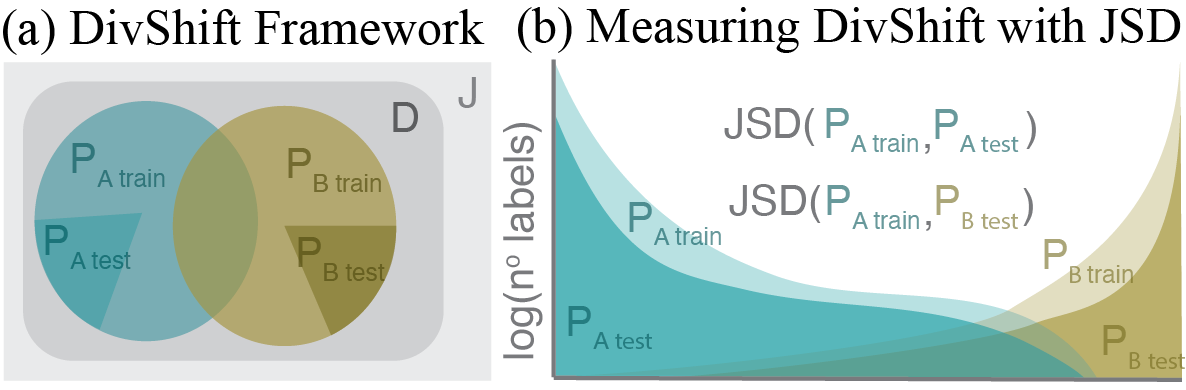}
        \end{center}
\caption{The Diversity Shift (DivShift) Framework \textbf{(a)} quantifies impacts of domain-specific biases by first partitioning data into partitions $P_A$ and $P_B$ using expert-verified types of bias. Bias impacts are then quantified by measuring the accuracy of models trained on $P_{A train}$ using $P_{A test}$ and $P_{B test}$ which is further compared to \textbf{(b)} the distribution shift between labels in $P_{A train}$ to labels in $P_{A test}$ and $P_{B test}$ using the Jensen-Shannon Distance (JSD).
\label{fig:f2}
}
\end{figure}

In order to quantify the performance effects of bias present in volunteer-collected biodiversity datasets, we propose Diversity Shift (DivShift), a new framework that casts these domain-specific biases as distribution shifts (Fig. ~\ref{fig:f2}). The DivShift framework quantifies the effect of bias by measuring the in-domain versus out-of-domain model performance of any two partitions of a dataset and further compares these changes to the underlying label distribution shift present across these two partitions. 

Given any finite labeled dataset $D$ consisting of pairs of inputs $x$ and labels $y$, we first define partition $P_A$ as any subset of $D$ such that $P_A \subset D$. We similarly define a second partition $P_B$ such that $P_B \subset D$ and $P_B \bigcap P_A = \emptyset$. These partitions are then each further split into two sub-partitions $P_{A train}$ and $P_{A test}$ where again $P_{A test} \bigcap P_{A train} = \emptyset$ (Fig. \ref{fig:f2}a). 
If the sampling process $\stackrel{S}{\sim}$ for $P_A$ and $P_B$ is identical, $P_A$ and $P_B$ are considered \emph{in-domain}. However, when the sampling processes for $P_A$ and $P_B$ are biased (e.g. more observers selectively uploading a few species in a certain area) then $P_A\stackrel{S_a}{\sim} J(x,y)$ and $P_B\stackrel{S_b}{\sim} J(x,y)$ will be out-of-distribution relative to each other, even if the underlying joint distribution $J(x,y)$---or the true representative biodiversity in an area---is the same. Therefore, any model trained on $P_{A train}$ will exhibit a changed performance on $P_{Btest}$ relative to $P_{A test}$ under these conditions.

To quantify this performance change, we first assume that the distribution of labels $y$ in $P_A$ and $P_B$ can be used to estimate their joint distributions and summarily use these labels to estimate the underlying distribution shift between these partitions. Namely, we measure the Jensen-Shannon Divergence (JSD) between $P_{A train}(y)$ and $P_{B test}(y)$, specifically using a base 2 log to ensure the distance is bound between $1$ and $0$ where $0$ is perfectly aligned and $1$ is perfectly misaligned distribution \cite{endres2003new} (Fig. \ref{fig:f2}b). We choose JSD for quantifying distribution shift as it is a bounded symmetric metric, which allows the comparison between changes in the magnitude of the JSD across partitions to changes in model performance. Furthermore, JSD is a well-established metric in the ecology literature for comparing biodiversity across sites. 

 While each $P_{train}$ and $P_{test}$ pair are uniformly sub-partitioned and sampled from the same distribution, the datasets are finite and the random sampling process is not truly random, meaning the label distributions between $P_{train}$ and $P_{test}$ will not perfectly match. To account for this noise in which observations are ultimately selected for each $P_{train}$, for each partition, the JSD between each paired $P_{train}$ and $P_{test}$ is subtracted from the estimates of distribution shift to other partitions. Specifically, the DivShift framework first measures the performance decrease between models trained on $P_{A train}$ and tested on $P_{B test}$ and compares those decreases to the JSD between $P_{A train}(y)$ and $P_{B test}(y)$ adjusted by the JSD between $P_{A train}(y)$ and $P_{A test}(y)$.

For any set of partitions where the underlying JSD is smaller than the difference in models' test accuracy across the partitions, we consider that to be a \emph{strongly biased} partition, implying that the distribution shift between $P_{A}(x,y)$ and $P_{B}(x,y)$ is even greater than the shift between $P_{A}(y)$ and $P_{B}(y)$. Conversely, partitions where the JSD is greater than the difference in model accuracy can be considered to be \emph{weakly biased} partitions, implying that the distribution shift between $P_{A}(x,y)$ and $P_{B}(x,y)$ is smaller than the shift between $P_{A}(y)$ and $P_{B}(y)$. 

While the magnitude of model performance change across partitions is informative for comparing to the underlying label distribution shift, to drill down on the importance of the sign of performance changes across partitions, the DivShift framework also measures the performance changes between models trained on $P_{A train}$ and $P_{B train}$ tested on both $P_{A test}$ and $P_{B test}$. Whether a model performs better or worse on its out-of-distribution test set partition depends on the nature of the biased samplers $ \stackrel{S_a}\sim$ and  $\stackrel{S_b}\sim$; or in other words, some biases in the data generation process may be more helpful than others for estimating the underlying distribution $J(x,y)$. Specifically, when a model trained on $P_{A train}$ has a higher out-of-partition accuracy on $P_{B test}$ than the model trained on $P_{B train}$, then the model trained on $P_{A train}$ is a \textit{strong generalizer} with respect to the model trained on $P_{B train}$, which is \textit{overfitted}. This implies that some structure in the joint distribution $P_{A}(x,y)$ captures useful information about $P_{B}(x,y)$ that $P_{B train}$ potentially lacks. 


\section{DivShift-NAWC Case Study}

\begin{figure*} [h]
    \begin{center}
        \includegraphics[width=0.95\linewidth]{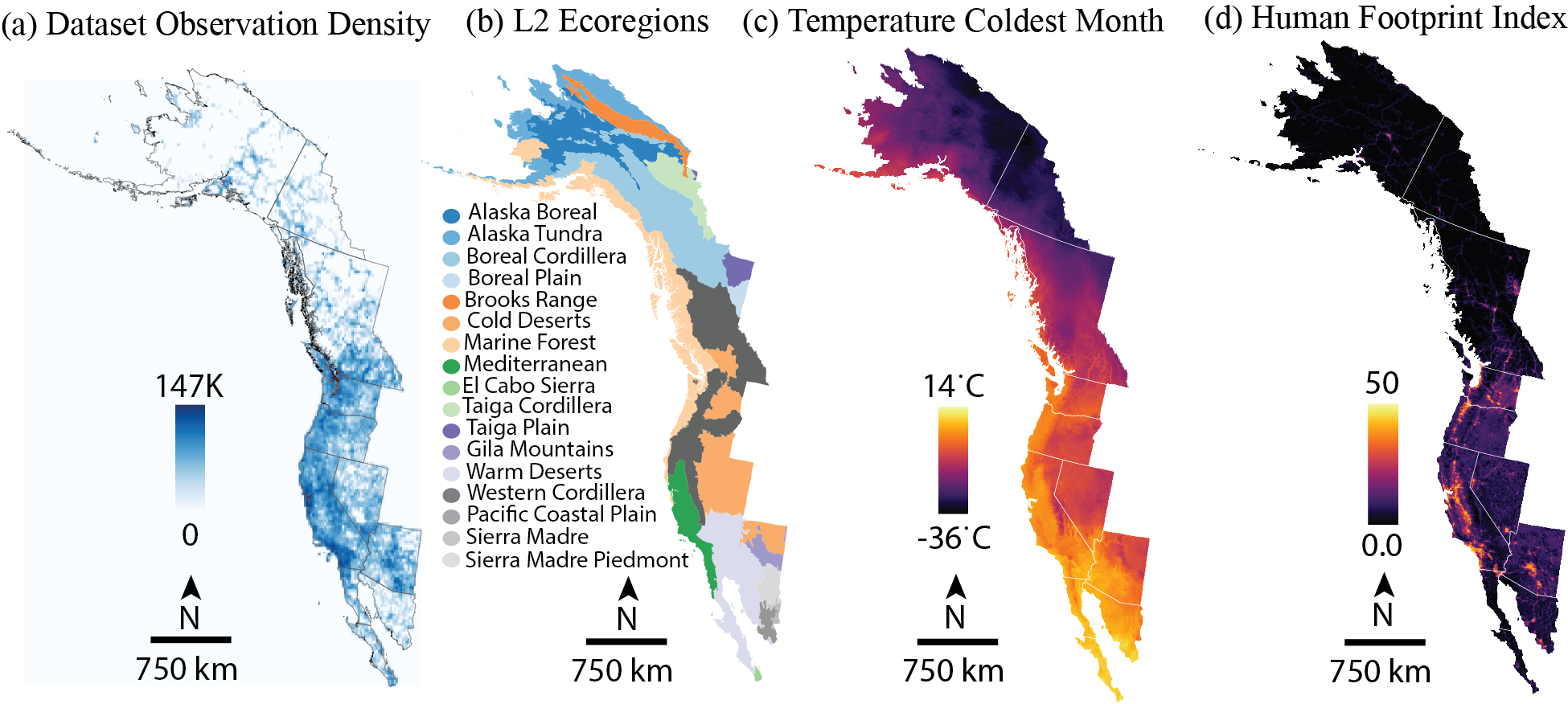}
    \end{center}
\caption{\textbf{Overview of DivShift--North American West Coast Dataset (DivShift-NAWC).} \textbf{(a)} Density plot of the DivShift-NAWC's iNaturalist observations \cite{inat}. Observations are skewed to U.S. and coastal states. \textbf{(b)} DivShift-NAWC spans a diverse set of habitats and ecosystems \cite{ecoregions}, \textbf{(c)} along with climates \cite{WorldClim}. \textbf{(d)} DivShift-NAWC observations are concentrated in human-modified areas \cite{HumanFootprint}.
\label{fig:f3}
}
\end{figure*}

To prototype the DivShift framework, we introduce the DivShift--North American West Coast (DivShift-NAWC) dataset. DivShift-NAWC consists of $\sim$7.3 million images from $\sim$3.9 million research-grade and in need of ID iNaturalist observations across the west coast of North America (Fig. \ref{fig:f3}). The states in DivShift-NAWC cover seven of the world’s nine terrestrial biomes, and include some of the coldest (Denali peak), hottest, driest (Death Valley), and wettest (Olympic peninsula) places on Earth. These states further include a high variation in socioeconomic status (2022 CA GDP: USD\$3,600 bil.; 2022 BS GDP: USD\$13 bil.) and data availability (CA: 6.22 obs/km\textsuperscript{2}; AK: 0.01 obs/km\textsuperscript{2}). 

\subsection{Distribution Shifts}

While there are many taxonomies for classifying the kinds of bias present in biodiversity data \cite{carlen2024framework, isaac2015bias}, for this work, we partition DivShift-NAWC based on five main types of bias: spatial, temporal, taxonomic, quality, and sociopolitical (Fig. \ref{fig:f1}). For each partition, we randomly split 80\% of the images into train and 20\% into test data.

To measure the underlying data partition distribution shifts, we filtered each paired partition to only species shared between the partitions, and used Scipy's JSD function with a log base of 2 to calculate the distance between the label distributions (Fig. \ref{fig:f2}b). To account for noise in the sampling process, for the JSD calculation we randomly resample the 80\% train 20\% test splits five times for each partition and report the mean and standard deviation. To measure the impact on machine learning model performance, we train models on both partitions and test on said partition plus its complement, comparing the differences in performance.

\begin{table}[h] 
\renewcommand{\arraystretch}{1.05} 
\small
\begin{tabularx}{\columnwidth}{|>{\centering\arraybackslash}p{3.15cm}>{\centering\arraybackslash}X>{\centering\arraybackslash}X>{\centering\arraybackslash}X>{\centering\arraybackslash}p{.6cm}|} 
  \hline
   \textbf{Partition} & \textbf{Images} & \textbf{Research Grade} & \textbf{Obs.} & \textbf{Spec.} \\ 
  \hline 
  
  DivShift-NAWC Dataset & 7.348M & 4.726M & 3.905M & 7,607\\ 
  \hline
  \multicolumn{5}{|c|}{\textbf{Baselines}} \\
  \hline
  iNat21
  & 3.554M & 3.554M & 1.937M & 1,852\\ 
  iNat21 mini
  & 0.185M & 0.185M & 0.109M & 1,852\\ 
  ImageNet
  & 1.614M & 1.614M & 0.858M & 1,260\\ 
  Spatial Stratified & 7.348M & 4.726M & 3.905M & 7,607\\ 
  \hline
  \multicolumn{5}{|c|}{\textbf{Taxonomic Bias}} \\
  \hline
  Long-tail
  & 4.725M & 4.725M & 2.527M & 7,607\\ 
  Balanced
  & 1.992M & 1.992M & 1.007M & 7,607\\ 
  \hline
  \multicolumn{5}{|c|}{\textbf{Temporal Bias}} \\
  \hline
  City Nature (CNC)
  & 0.362M & 0.245M & 0.220M & 3,929\\ 
  Not City Nature 
  & 6.986M & 4.480M & 3.685M & 7,604\\ 
  \hline
  \multicolumn{5}{|c|}{\textbf{Observer Bias}} \\
  \hline
  Engaged
  & 3.476M & 2.324M & 1.697M & 7,361\\ 
  Casual 
  & 1.113M & 0.660M & 0.756M & 5,706\\ 
  \hline
  \multicolumn{5}{|c|}{\textbf{Spatial Bias}} \\
  \hline
  Modified
  & 6.642M & 4.280M & 3.536M & 7,513\\ 
  Wilderness
  & 0.141M & 0.083M & 0.068M & 2,395\\ 
  \hline
  \multicolumn{5}{|c|}{\textbf{Sociopolitical Bias}} \\
  \hline
  Alaska (AK)
  & 0.099M & 0.064M & 0.057M & 875\\ 
  Arizona (AZ)
  & 0.497M & 0.313M & 0.272M & 2,191\\ 
  Baja California (BN)
  & 0.142M & 0.098M & 0.090M & 1,466\\ 
  Baja California Sur (BS)
  & 0.046M & 0.033M & 0.022M & 716\\ 
  British Columbia (BC)
  & 1.080M & 0.691M & 0.622M & 2,329\\ 
  California (CA)
  & 4.039M & 2.558M & 2.115M & 4,654\\ 
  Nevada (NV)
  & 0.259M & 0.177M & 0.121M & 1,860\\ 
  Oregon (OR)
  & 0.604M & 0.399M & 0.300M & 2,711\\ 
  Sonora (SO)
  & 0.018M & 0.010M & 0.010M & 673\\ 
  Washington (WA)
  & 0.529M & 0.357M & 0.279M & 2,393\\ 
  Yukon (YK)
  & 0.034M & 0.026M & 0.018M & 746\\ 
  \hline
\end{tabularx}
\caption{DivShift-NAWC Data Representation by Partition. Obs=Observations. Spec=Species. Research-grade images are verified by at least two iNaturalist community members.}
\label{table:Splits}
\end{table}

\subsubsection{Spatial Partition: Human Footprint}
Human-driven land use change is widespread across the planet, but there still exist large tracts of undisturbed habitat especially in the polar regions (Fig. \ref{fig:f3}d). However, these wilder regions are also harder to reach, making it difficult for volunteers to collect imagery there (Fig. \ref{fig:f1}a) and skewing volunteer-collected biodiversity data towards human-modified habitats (Fig. \ref{fig:f4}a). Using the Global Human Footprint Index (HFI) ~\cite{HumanFootprint}, we partition DivShift-NAWC into wilderness (HFI $\leq$ 1) and highly modified observations (HFI $\geq$ 4). Interestingly, over 90\% of the 7.3 million images in DivShift-NAWC are from highly human-modified regions while only $\sim$6\% are from minimally-modified wilderness (Table \ref{table:Splits}), as compared to $\sim$48\% of all landmass in the DivShift-NAWC states being wilderness versus $\sim$37\% being highly-modified. 

\subsubsection{Temporal Partition: City Nature Challenge}
\begin{figure*}[h]
        \begin{center}
                \includegraphics[width=\linewidth]{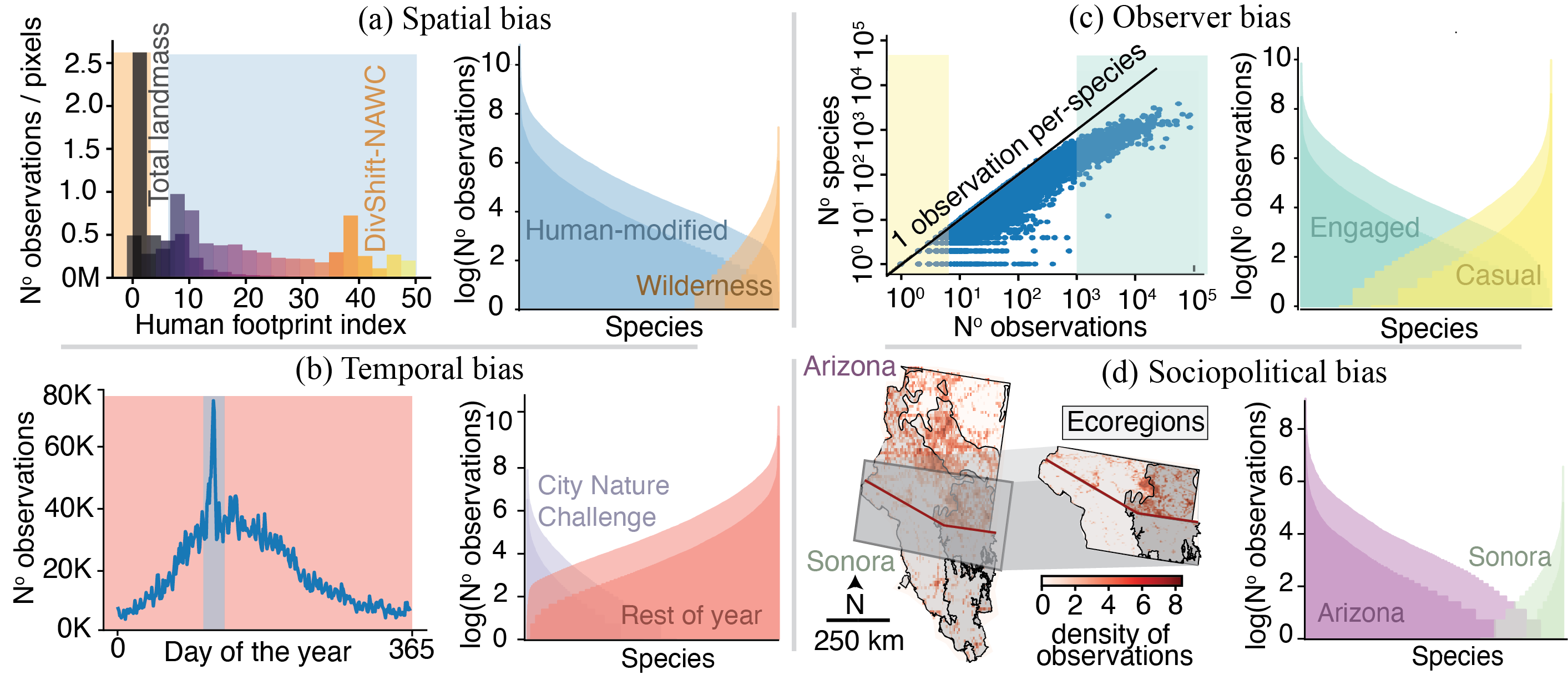}
        \end{center}
\caption{Biases in the DivShift-NAWC dataset. \textbf{(a)} Human footprint index \cite{HumanFootprint} across human-modified and wilderness areas. \textbf{(b)} Observations per-day, with City Nature Challenge spike highlighted. \textbf{(c).} Observations per-observer with casual/engaged lines highlighted. \textbf{(d)} Density of observations in shared ecoregions across Arizona-Sonora border.
\label{fig:f4}
}
\end{figure*}

The City Nature Challenge happens every year during the last weekend in April. This challenge creates a large spike in observations (Fig. \ref{fig:f4}b) ~\cite{di2021observing, inat_review} and leads to altered observer behavior (Fig. \ref{fig:f1}b), as volunteers are encouraged to maximize the number of observations and unique species they observe within the week. While the majority of iNaturalist photos are taken outside of this challenge, a higher proportion of observations from the City Nature Challenge are labeled. Indeed, the Challenge captures more than half of the species from the entire DivShift-NAWC dataset despite having less than 6\% of the total observations (Table \ref{table:Splits}), implying that observer behavior patterns shift significantly during the challenge. To test the benefits and drawbacks of this altered user behavior on model performance, we partition DivShift-NAWC so all observations taken during official City Nature Challenge (CNC) dates for the four years of study comprise one partition, while observations from all other weeks comprise the other.

\subsubsection{Taxonomic Partition: Long-Tailed Versus Balanced}
While most species are rare and few species are common \cite{enquist2019commonness}, volunteer observations tend to be especially skewed towards charismatic or interesting species (Figs. \ref{fig:f1}c, \ref{fig:taxanomic-intro}a). To minimize these long-tail performance effects, many computer vision datasets built from volunteer-collected biodiversity image collections tend to re-balance the number of examples per-species to be more uniform (Fig. \ref{fig:taxanomic-intro}b) \cite{iNat2021, deng2009imagenet}. However, some of these more commonly observed species are also more ecologically abundant, and thus have a larger diversity of phenotypes, leading to a larger intra-class variability. This artificial balancing decision thus means that additional observations for these abundant species are excluded, potentially harming performance for these common species. 

To explore the ramifications of this choice, we compare the performance of models trained on two train partitions with the same test set. Namely, after removing species with fewer than 25 observations, we consider a long-tailed partition where all observations are kept for common species. We also consider a balanced partition, where we only train on up to 300 randomly-selected images for common classes, ultimately discarding observations for 2,375 of the 7,607 species in this partition.

\begin{table}[h]
\renewcommand{\arraystretch}{1.05} 
\small


\begin{tabularx}{\columnwidth}{|>{\centering\arraybackslash}
p{1.5cm}
>{\centering\arraybackslash}
p{1.4cm}>{\centering\arraybackslash}p{2cm}>{\centering\arraybackslash}X>{\centering\arraybackslash}X|}
\hline
\textbf{Train Partition} & \textbf{Test Partition} & \textbf{JSD Diff $(\mu \pm \sigma)\times100$} & \textbf{Top1-Img Diff}  & \textbf{Top1-Spec Diff} \\ \hline     
       \multicolumn{5}{|c|}{\textbf{Spatial Bias (Human Footprint)}} \\ \hline
        Wild & Modified  & 49.29 $\pm$ 0.15 & \textcolor{red}{-35.3} & \textcolor{red}{-15.4} \\ 
        Modified& Wild  & 71.87 $\pm$ 0.30 & \textcolor{red}{-11.3} & \textcolor{green}{+6.9} \\ 
       \hline
       \multicolumn{5}{|c|}{\textbf{Temporal Bias (City Nature Challenge (CNC))}} \\ \hline
         CNC & Not CNC & 19.53 $\pm$ 0.08 & \textcolor{red}{-17.9} & \textcolor{red}{-10.0}\\ 
         Not CNC&CNC  & 36.39 $\pm$ 0.14 & \textcolor{green}{+1.9} & \textcolor{green}{+9.2} \\ 
       \hline
       \multicolumn{5}{|c|}{\textbf{Observer Bias (Observer Engagement)}} \\ \hline
        Casual& Engaged & 26.70 $\pm$ 0.09 & \textcolor{red}{-23.1}& \textcolor{red}{-12.5} \\ 
        Engaged&Casual & 33.20 $\pm$ 0.08 & \textcolor{green}{+4.9}& \textcolor{green}{+3.8} \\ 
        \hline
       \multicolumn{5}{|c|}{\textbf{Sociopolitical Bias (State Borders)}} \\ \hline
        CA&BC & 79.04 $\pm$ 0.08 & \textcolor{red}{-29.1}& \textcolor{red}{-15.2}\\ 
        CA&WA & 77.08 $\pm$ 0.09& \textcolor{red}{-26.1}& \textcolor{red}{-11.6}\\ 
        CA&OR & 70.27 $\pm$ 0.06& \textcolor{red}{-24.6}& \textcolor{red}{-0.9}\\ 
        CA&AZ & 74.83 $\pm$ 0.09& \textcolor{red}{-23.0}& \textcolor{red}{-6.5}\\ 
        CA&NV & 75.64 $\pm$ 0.06& \textcolor{red}{-14.2}& \textcolor{red}{-0.2}\\ 
        CA&BN & 59.66 $\pm$ 0.13& \textcolor{red}{-10.3}& \textcolor{green}{+5.9}\\ 
        CA&BS & 87.52 $\pm$ 0.15& \textcolor{red}{-39.9}& \textcolor{red}{-10.1}\\ 
        CA&SO & 86.77 $\pm$ 0.39& \textcolor{red}{-26.1}& \textcolor{green}{+2.9}\\ 
        BC&AK  & 65.28 $\pm$ 0.26 & \textcolor{red}{-20.5}& \textcolor{red}{-3.7} \\ 
        BC&YK  & 75.61 $\pm$ 0.17& \textcolor{red}{-32.4}& \textcolor{red}{-10.8}\\ 
        BC&WA  & 28.36 $\pm$ 0.18& \textcolor{red}{-8.9}& \textcolor{red}{-5.4}\\ 
        BC&OR  & 38.72 $\pm$ 0.13& \textcolor{red}{-16.7}& \textcolor{red}{-7.0}\\ 
        BC&CA  & 62.38 $\pm$ 0.12 & \textcolor{red}{-29.1}& \textcolor{red}{-8.5}\\  \hline
\end{tabularx}
\caption{Comparison of label distribution shift to performance shift across bias partitions on DivShift-NAWC. The difference in JSD is calculated as $JSD(P_{A train}, P_{B test}) - JSD(P_{A train}, P_{A test})$ and the difference in  observation performance (in $\%$) is the model's out-of-distribution's test set performance minus the test set performance on the in-distribution train partition. JSD=Jensen-Shannon Divergence, Diff=Difference, Img=Image, Spec=Species.}
\label{table:jsd-results}
\end{table}

\subsubsection{Observer Partition: Observer Engagement}
Since iNaturalist observations are collected by volunteers with differing amounts of enthusiasm, time, and resources \cite{mac2020citizens, blake2020demographics}, observer engagement varies widely between observers (Fig. \ref{fig:f1}d). Given that observers who use the app frequently and collect more data tend to observe a wider diversity of species in more diverse habitats (Fig. \ref{fig:f4}c) \cite{di2021observing}, we also partition DivShift-NAWC by user engagement, with the casual partition consisting of all observations from observers with fewer than 50 total research-grade observations, and the engaged partition as observations from observers with more than 1,000 research-grade observations \cite{di2021observing}.


\subsubsection{Sociopolitical Partition: State Boundaries}
Where certain plant species can grow are demarcated by ecological boundaries (Fig. \ref{fig:f3}b). Similarly, volunteer observation trends are demarcated by political boundaries which may not necessarily reflect ecological ones (Figs. \ref{fig:f1}e). For example, while the Sonora and Mojave deserts extend beyond the borders of California and Arizona, the stark effects of political boundaries can be seen in the difference in abundance of observations between the U.S. and Mexico, especially across the Arizona--Sonora border, which bifurcates two similar ecosystems (Fig. \ref{fig:f4}d). 

To test whether predictive accuracy of models trained in observation-rich geographies can extend across these at times ecologically arbitrary political boundaries to observation-poor regions, we compare model performance trained with observations from two states with a large number of observations (British Columbia and California) and test them on nearby states with varying levels of volunteer-collected biodiversity data availability (Alaska, Washington, Oregon, Yukon, and California for British Columbia; British Columbia, Washington, Oregon, Arizona, Nevada, Baja California, and Baja California Sur for California).

\begin{table}[h]
\renewcommand{\arraystretch}{1.05} 
\small
\begin{tabularx}{\linewidth}{|>{\centering\arraybackslash}p{2.5cm}>{\centering\arraybackslash}X>{\centering\arraybackslash}X>{\centering\arraybackslash}X>{\centering\arraybackslash}X>{\centering\arraybackslash}X>{\centering\arraybackslash}X|} 
\hline
\textbf{Train-Test}  & \textbf{Wgt}  & \textbf{FAR}  & \textbf{CAR}  & \textbf{RAR } & \textbf{LUC } & \textbf{Eco }\\ \hline     
       \multicolumn{7}{|c|}{\textbf{Spatial Bias (Human Footprint)}} \\ \hline
        Wild-Wild & 16.1 & 34.8 & 30.0 & 18.1 & 55.5 & 52.8 \\ 
        Modified-Wild   & \textbf{43.2} & \textbf{66.5} & \textbf{44.0} & 10.8 & \textbf{61.9} &  \textbf{59.5}\\ 
        Modified-Modified  & 15.6 & 69.3 & 48.8 & 21.3 & 70.2 & 70.1\\ 
        Wild-Modified  &  8.5&  19.2&  12.4&  5.2&  17.9& 18.0  \\ 
       \hline
       \multicolumn{7}{|c|}{\textbf{Temporal Bias (City Nature Challenge (CNC))}} \\ \hline
         CNC-CNC  & 8.7 & 32.5 & 13.8 & 5.2 & 43.4 & 49.0 \\ 
         Not CNC-CNC   & \textbf{37.0} & \textbf{65.8} & \textbf{44.7} & \textbf{18.6} & \textbf{70.8} & \textbf{75.8} \\ 
         Not CNC-Not CNC  & 17.6 & 69.0 & 49.2 & 22.7 & 71.1 & 69.5 \\ 
         CNC-Not CNC  & 2.2 & 21.8 & 6.1 & 1.4 & 27.0 & 26.2 \\ 
       \hline
       \multicolumn{7}{|c|}{\textbf{Taxonomic Bias (Balanced vs Long Tailed)}} \\ \hline
        Long-Long & 17.8 & 70.9 & 50.4 & 24.7 & 70.1 & 72.0 \\ 
        Balanced-Long & \textbf{19.2} & 55.0 & \textbf{52.5} & \textbf{26.6} & 52.0 & 51.3 \\ 
       \hline
       \multicolumn{7}{|c|}{\textbf{Observer Bias (Observer Engagement)}} \\ \hline
        Casual-Casual & 11.4& 51.6& 22.3& 8.7& 61.0 & 63.8 \\ 
        Engaged-Casual  & \textbf{24.6} & \textbf{62.5} & \textbf{36.7} & \textbf{11.1} & \textbf{65.9} & \textbf{70.6}\\ 
        Engaged-Engaged & 15.0& 63.2& 41.2& 18.4& 63.0 & 62.1 \\ 
        Casual-Engaged  & 4.7 & 39.9 & 11.2 & 1.8 & 41.8 & 41.3 \\ 
        \hline
       \multicolumn{7}{|c|}{\textbf{Sociopolitical Bias (State Borders)}} \\ \hline
        CA-CA  & 14.6& 63.9& 44.6& 21.0& 72.0 & - \\ 
        CA-BC  & 20.6& 42.5& 17.0& 4.6& 41.1 & -\\ 
        CA-WA  & 24.2& 47.6& 21.2& 3.9& 45.7 &-\\ 
        CA-OR  & 26.6& 49.9& 27.6& 7.6& 47.2 &-\\ 
        CA-AZ  & 29.3& 51.5& 25.3& 8.7& 50.1 &-\\ 
        CA-NV  & 39.0& 56.1& 35.3& 19.9& 56.1 &-\\ 
        CA-BN  & 36.2& 62.3& 37.7& 11.3& 60.1 &-\\ 
        
        CA-BS  & 32.2& 49.7 & 23.3& 4.2& 32.8 & -\\ 
        CA-SO  & 38.3 & 50.3& 22.0  & 11.4& 51.6 & -\\ 
        BC-BC & 11.4& 57.5& 35.0& 13.6& 70.5 &-\\ 
        BC-AK  & 21.4& 55.6& 23.4& 4.4& 47.8 &-\\ 
        BC-YK & 17.9& 49.8& 23.3& 2.3 & 34.9 & -\\ 
        BC-WA  & 12.5& 50.8& 23.1& 3.6& 59.3 &-\\ 
        BC-OR  & 15.3& 46.3& 20.0& 4.4& 50.6 &-\\ 
        BC-CA  & 18.2& 43.5& 17.2& 1.7& 40.7 &-\\  
        \hline
\end{tabularx}
\caption{Top-1 accuracy results (in \%) on DivShift-NAWC across bias partitions. Wgt=Weighted, FAR=Frequent Average Recall, CAR=Common Average Recall, RAR=Rare Average Recall, LUC=Land Use Category. Top-1 Eco is excluded for sociopolitical bias due to the lack of consistency of overlapping ecoregions between states.}
\label{table:results}
\end{table}

\subsubsection{Baseline Partitions}
Lastly, we compare the absolute accuracy of these partitions to a variety of classic partitioning schemes from natural world imagery datasets. Specifically, we recreated the filtering and partitioning schema of the iNat2021 benchmarking dataset \cite{iNat2021}. 
We also recreated the iNat2021 mini train partition by randomly sub-sampling exactly 50 images per-species from the train set. Additionally, we tested spatial stratification, partitioning the study area into a 50 x 50 km grid and randomly assigning 20\% of the grid boxes and the DivShift-NAWC images that fell within these boxes to test and the rest to train \cite{cole2020geolifeclef, crisp}. We also recreated the Imagenet train / test partitioning strategy, \cite{deng2009imagenet}
and tested a naive train / test partition where 20\% of observations and all of their corresponding images are selected as test images while the rest are used for training. 

\subsection{Model Training and Testing}
To quantify bias effects with the DivShift-NAWC dataset, we focus on the specific task of fine-grained species recognition (identifying species from their images), which is an important biodiversity monitoring task for automating species detection. We only use research-grade iNaturalist observations as these images have been community-verified that the species identification is correct. As the goal of this work is to test distribution shift effects across partitions of volunteer-collected data as opposed to maximizing predictive performance, for each partition we train a small computer vision model for a limited number of epochs with the same hyperparameter configuration for each model. 

Specifically, for each partition we train a ResNet-18 initialized with ImageNet pre-trained weights for 10 epochs with a batch size of 64, an SGD optimizer, single-label cross-entropy loss, and a learning rate of 0.064. Image augmentations were limited to resizing each image to at least 256 x 256 pixels and center cropping to 224 pixels, then normalizing the image with Imagenet mean and standard deviation. For testing, we employ early stopping using Top-1 species accuracy, and for all partitions we test only with images from species present in the split the model was trained on. To demonstrate how architecture and size choices improve absolute accuracy, we also ablate the model architecture and size, training a large ResNet50 and a base-size vision transformer (ViT) on the casual partition of the dataset.

Models' accuracies were measured using eight diverse accuracy metrics common to the machine community, such as Top-1 per-image (Top1-Img), per-species accuracy (Top1-Spec), accuracy aggregated by rarity \cite{AutoArborist}(Top1-FAR, CAR, RAR), and accuracy broken down by ecoregion (Top1-Eco) \cite{crisp}. We also introduce a new rarity-weighted loss function that emphasizes performance on classes that are rare within a partition (Top1-Wgt), and accuracy broken down by land use category (Top1-LUC), emphasizing performance across both human-modified and untouched habitats.


\section{Results and Discussion}
Generally speaking, comparing the dataset label distribution shift measured by the JSD (Table \ref{table:jsd-results}, JSD Diff) versus the model performance shift (Table \ref{table:jsd-results},  Top1-Img Diff), when looking at strong versus weak biases per partition, we see encouragingly that all bias partitions are weakly biased as overall model performance changes  for in- vs. out-of-distribution are smaller than the JSD across labels. For the sociopolitical bias partition, despite having the largest in- vs. out-of-domain model performance drops (Table \ref{table:jsd-results}, Sociopolitical Bias, Top1-Img), these drops in performance are much smaller than the drop in JSD between the partitions (Table \ref{table:jsd-results}, Sociopolitical Bias, JSD Diff). This implies that computer vision model performance losses across geographies tend to be less pronounced than underlying data availability differences \cite{beery2018recognition}.

\subsubsection{Wilderness Vs. Modified Habitats} For the spatial split, training on observations from less-disturbed habitats (Table \ref{table:results}, Wild-Modified) leads to worse performance than training in areas of high human activity (Table \ref{table:results}, Modified-Wild), likely due to the significant difference in the number of observations between these partitions, with modified regions having more than 3.5 million more unique observations than wilderness regions (Table \ref{table:Splits}, Spatial Bias). 
Indeed, the modified-trained model is a strong generalizer to wilderness regions, showing a higher out-of-distribution accuracy on wilderness observations for all but the rarest species in the DivShift-NAWC dataset (bolded entries, Table \ref{table:results}, Modified-Wild; Fig. \ref{fig:spatial}a). The wilderness-trained model's low transferability even for rare species implies it has overfit and the wilderness partition is strongly biased, lacking sufficient data to build well-generalized models. 

Further evidence of the paucity of data from wilderness regions include the fact that the modified-trained model's out-of-distribution rarity-weighted accuracy for rare species within the wilderness partition (Wgt, Table \ref{table:results}, Modified-Wild) is much higher than its accuracy on species rare across the entire DivShift-NAWC dataset (RAR, Table \ref{table:results}, Modified-Wild). This implies that many species that are rare in wilderness regions are in fact commonly dispersed in highly modified regions. Alternately, we see that for species most commonly observed in wilderness regions, the wilderness-trained model outperforms the modified-trained model, but only for observations for those species taken in wilderness regions (Fig. \ref{fig:spatial}b). This suggests that the growth form or environment of these wilderness–associated species is highly unique, further underscoring the need for more data collection across our wildest landscapes.

\subsubsection{City Nature Challenge}  For the temporal split, we also see uniformly worse performance training on observations from the City Nature Challenge and testing on observations from outside the Challenge (Table \ref{table:results}, CNC-Not CNC; Fig. \ref{fig:temporal}a),  likely a consequence of significant differences in partition size (Table \ref{table:Splits}, Temporal Bias).
Meanwhile, training on observations from outside the Challenge strongly generalizes and leads to improved performance on observations from the Challenge in all cases  (bolded entries, Table \ref{table:results}, Not CNC-CNC), while the Challenge-trained model is especially overfitted for rare species and species-rich portions of the year outside of the City Nature Challenge (Table \ref{table:results}, CNC-Not CNC, RAR; Fig. \ref{fig:temporal}b). This implies that during the challenge iNaturalist users are more drawn to increase their species count as opposed to their observation count during this bioblitz, leading to too many species and not enough observations to effectively train models from the City Nature Challenge ($\sim62$ observations per-species CNC, $\sim589$ observations per-species Not CNC).
\begin{table*}[t]
\renewcommand{\arraystretch}{1.05} 
\small
\begin{tabularx}{\linewidth}{|>{\centering\arraybackslash}Y>{\centering\arraybackslash}p{2cm}>{\centering\arraybackslash}X>{\centering\arraybackslash}X>{\centering\arraybackslash}X>{\centering\arraybackslash}X>{\centering\arraybackslash}X>{\centering\arraybackslash}X>{\centering\arraybackslash}X>{\centering\arraybackslash}X|} 
\hline
\textbf{Baseline}  & \textbf{JSD $(\mu \pm \sigma)\times100$} & \textbf{Top1-Img \%}  & \textbf{Top1-Spec \%} & \textbf{Top1-Wgt \%}  & \textbf{Top1-FAR \%}  & \textbf{Top1-CAR \%}  & \textbf{Top1-RAR \%} & \textbf{Top1-LUC \%} & \textbf{Top1-Eco \%}\\ \hline
        Spatial &31.23 & 66.3 &37.6 &21.4& 65.8 & 42.6 & 15.6 & 65.1 & 66.1 \\ 
        ImageNet &7.27 & 69.5 & 69.5 & 69.5 & 69.5 & - & - & 69.9 & 70.4 \\ 
        iNat2021 &40.20 & 68.0 & 68.0 & 68.0 & 71.8 & 52.5 & - & 69.1 & 64.7 \\ 
        iNat2021 Mini &0.00 & 33.7 & 33.7 & 33.7 & 34.0 & 32.3 & - &31.0 & 29.3 \\
        Random &8.09 & 70.6 & 40.5 & 20.5 & 69.8 & 48.4 & 21.3 & 70.9 & 69.8 \\  \hline
\end{tabularx}
\caption{Results for baseline partitions \cite{crisp, deng2009imagenet,  iNat2021}. Some CAR and RAR values are missing due to lack of common and rare species in baseline partition.}
\label{table:baseline}
\end{table*}

\begin{table}[b]
\renewcommand{\arraystretch}{1.05} 
\small
\begin{tabularx}{\linewidth}{|>{\centering\arraybackslash}Y|>{\centering\arraybackslash}X>{\centering\arraybackslash}Z|>{\centering\arraybackslash}X>{\centering\arraybackslash}Y|} 
\hline
\textbf{Metric} & \textbf{ResNet50} & \textbf{ResNet50 Diff} & \textbf{ViT} & \textbf{ViT Diff} \\ \hline
\textbf{Img \%} & 71.1 & -22.7 & 79.0 & -23.1 \\ 
\textbf{Spec \%}  & 35.2 & -14.1 & 47.4 & -21.0 \\ 
\textbf{Wgt \%} & 15.6 & -8.1 & 28.0 & -17.6 \\ 
\textbf{FAR \%} & 59.1 & -11.4 & 70.2 & -13.9 \\ 
\textbf{CAR \%} & 28.4 & -12.9 & 42.3 & -20.7 \\ 
\textbf{RAR \%} & 13.0 & -9.4 & 24.0 & -19.0 \\ 
\textbf{LUC \%} & 74.0 & -26.6 &  78.5 & -21.5 \\ 
\textbf{Eco \%} & 65.6 & -16.4 & 81.1 & -24.4 \\ \hline
\end{tabularx}
\caption{Top-1 in-distribution and out-of-distribution difference accuracy results for ablated model architecture and size. Models trained on Casual Observers and tested on Casual Observers and Engaged Observers.}
\label{table:extended}
\end{table}

\subsubsection{Long-Tailed Vs. Balanced}
For the taxonomic partition, we see that as expected, sub-sampling the most frequent classes improves accuracy per-class for all but the most common species (Table \ref{table:results}, Balanced-Long, FAR; Fig. \ref{fig:taxanomic} All Species). Indeed, balancing the training set leads to strong generalization for rare species (bolded entries, Table \ref{table:results}, Balanced-Long; Fig. \ref{fig:taxanomic}, By Species Rarity, Rare), but conversely using the maximal training data available leads to the best common species performance (Table \ref{table:results}, Long-Long, FAR; Fig. \ref{fig:taxanomic}, By Species Rarity, Frequent), highlighting the inherent tension between maximizing rare vs. common species performance.

\subsubsection{Casual Vs. Engaged Observers} 
 The observation quality split shows the most marked performance differences across partitions, with the model trained on observations from engaged users showing substantially better and strongly generalized performance
 (bolded entries, Table \ref{table:results}, Engaged-Casual) over models trained with observations from casual users (Table \ref{table:results}, Casual-Casual). Given that this bias partition has the most similar number of observations between the two partitions, this implies that the overall lower performance of the casual-trained model stems from lower image quality for observations taken by casual users as opposed to insufficient training data volumes. Indeed, if we bin performance by observer experience, we see that the engage-trained model has consistent and significantly higher generalization performance than the casual-trained model (Fig. \ref{fig:quality}).

Interestingly, for the engaged-trained model we see a strong negative relationship between the number of observations a user has uploaded and model performance (Fig. \ref{fig:quality}, engaged trained), suggesting that iNaturalist users who use the app least frequently generate more generic and easy to identify data. Conversely, the casual-trained model shows a strong positive relationship between user observations and performance (Fig. \ref{fig:quality}, casual trained), suggesting that  the most engaged users take the best photos for species identification and know what features and phenotypes to focus so individuals are easily identifiable.

\subsubsection{State Boundaries} For the sociopolitical boundaries, we see substantial distribution shifts between states, with a general correlation between distance in space and distance in distribution (Table \ref{table:jsd-results}, Sociopolitical Bias, JSD Diff). 
Similarly, model accuracy drops off with larger distances between states and fewer shared ecoregions (Fig. \ref{fig:socio-political}). However, when controlling for distance, states with a much higher data density tend to have a much lower performance (Table \ref{table:results}, Sociopolitical Bias, CA-BC, BC-CA) than similarly distant states with a low data density (Table \ref{table:results}, Sociopolitical Bias, CA-BS, BC-AK), implying that the data volume of low density states may be insufficient to reliably test model transfer performance.
These results imply that while predictive power decreases across boundaries, there is still some transferability across geography in the North American West Coast when data density is sufficient.

\subsubsection{Baselines}
For the baseline partitions, we find that in general performance is on par with that of the bias partitions trained on their in-domain test sets, like the engaged and long-tailed partitions (Table \ref{table:baseline}). While the Random partition has high per-image accuracy, its much lower rarity-weighted accuracy implies that much of these gains may be concentrated in just the most common species, a common critique of this sampling approach. The iNat2021 Mini split has the lowest absolute but also most consistent performance across accuracy metrics,  because sub-sampling a small number of images for every dense-enough class likely captures
a less-biased sub-sampling in expectation. Lastly, the spatial split has frequency-binned accuracies comparable to other baseline splits but a significantly lower species- and rarity-weighted accuracy, implying in aggregate that spatial block sampling can capture dataset-wide trends, but sometimes at the cost of high train-test variance (JSD: 31.23, Table \ref{table:baseline}).

\subsubsection{Model Architecture and Size Ablation}
 Comparing performance across the casual split of the quality bias partition for a larger ResNet architecture and a transformer-based architecture (Table \ref{table:extended}), we see both ablations outperform the ResNet-18 model across all accuracy metrics. Using a larger ResNet model led to modest performance improvements between $\sim$2 and $\sim$12\% depending on the metric. Using a more modern vision transformer architecture led to significantly larger performance improvements, between $\sim$15 and $\sim$20\% depending on the metric. Importantly, the differences between in-domain and out-domain per-image accuracy stay relatively consistent between ablations (Table \ref{table:jsd-results}, Casual-Engaged, Top1-Img Diff; Table \ref{table:extended}, Img \% Diff), highlighting the effectiveness of the DivShift framework to measure performance shifts independent of modeling choice.
\subsection{Recommendations for Downstream Modeling}
\textbf{Spatial Bias Takeaways:} Wilderness regions simply lack sufficient volunteer-collected biodiversity data to train effective models (Table \ref{table:jsd-results}, Human Footprint; Fig. \ref{fig:spatial}), thus downstream biodiversity modeling efforts targeted for undisturbed regions and their species will likely require additional data collection. 
\textbf{Temporal Bias Takeaways:} Normal iNaturalist user behavior leads to denser training data than from the City Nature Challenge data collection campaign alone (Table \ref{table:results}, CNC; Fig. \ref{fig:temporal}), thus modelers should complement models trained on bioblitz observations with data taken from the rest of the year when possible.
\textbf{Taxonomic Bias Takeaways:} Using more data even if long-tailed improves common species performance but reduces rare species performance ; Fig. \ref{fig:taxanomic},
 leading to an inevitable rare vs. common trade-off
(Table \ref{table:results}, Balanced vs Long-Tailed). 
Thus, training data sub-sampling should be chosen with downstream  biodiversity monitoring use cases in mind (e.g. maximal accuracy for detection of common invasive species, vs. endangered species recognition). 
\textbf{Observer Bias Takeaways:} Observations from more engaged observers are of resounding higher quality (Table \ref{table:results}, User Engagement; Fig. \ref{fig:quality}), thus modelers should consider discarding observations from observers with $<50$ observations.
\textbf{Sociopolitical Bias Takeaways:} Accuracy across geographies tends to degrade with larger distances but is obscured when data density is low  (Table \ref{table:jsd-results}, State Borders; Fig. \ref{fig:socio-political}), thus modelers working in data-sparse regions should take care to validate models with expert-collected data when possible. 


\subsection{Limitations and Future Work}
Our findings represent the first comprehensive effort to quantify and document the downstream effects of bias in biodiversity data on computer vision model species recognition performance. Despite documenting the effects of five unique partitions, there are yet even more kinds of biases not tested here, and further complex interactions and intersections between these biases should be explored \cite{carlen2024framework, bowler2022temporal}. So long as those biases enable the partitioning of biodiversity datasets, the flexibility of the DivShift framework should allow for the targeted testing of these additional and intersectional biases. 

While our framework can provide quantitative estimates of underlying distribution shift, it still lacks a mechanism to causally attribute performance changes to a given bias, an important direction for future work.
Similarly, we only evaluate common supervised approaches to contrast performance across shifts. In future work, we envision expanding DivShift to the unsupervised and long-tailed learning settings to benchmark more modern machine learning techniques for dealing with distribution shift.

Furthermore, relying on label distribution shift across iNaturalist images may not be the most biologically-plausible way to measure ecological shifts, and alone may not capture all correlations between features and labels influenced by environmental or other factors. In the future, we aim to measure underlying distribution shift within environmental space by comparing climate hulls across the bias partitions, and within image space by comparing shifts within image features via image embedding manifold analysis.

Lastly, by performing train/test splitting per-image instead of per-observation, within partitions there is the risk that different images from the same observation end up in both the train and test split, inflating in-distribution test performance and reducing model generalizability. The DivShift framework can directly test for these effects when models show high in-distribution but low out-of-distribution performance. Furthermore, the DivShift framework is agnostic to train/test splitting choices and future versions of the DivShift-NAWC dataset will include both per-observation and per-image splitting options.

\subsection{Conclusion}
Here we present DivShift, a framework for quantifying bias-induced distribution shift across biodiversity datasets and introduce DivShift-NAWC, a new large-scale natural world imagery dataset designed to benchmark distribution shift effects on computer vision model performance for biodiversity monitoring tasks like fine-grained species recognition. 
This framework and dataset enable the rigorous testing of problems known to the conservation biology community in a machine learning setting to help enable the building of more robust, accurate biodiversity monitoring tools from large-scale volunteer datasets.

\section{Acknowledgments}
We first thank all the participants who contributed observations on iNaturalist. We further thank Noah Goodman and Gabriel Poesia for their comments and discussion, and we further thank CoCoLab for donating compute for this project. We also thank the TomKat Center for Sustainable Energy, The Fulbright Brasil Commission, Carnegie Science, the Howard Hughes Medical Institute, and the University of California, Berkeley for funding support for this research. This research was also funded by the NSF Graduate Research Fellowship DGE-1656518 (L.G.), the TomKat Graduate Fellowship for Translational Research (L.G.), and the Krupp Internship Program for Stanford Students in Germany (E.S.). T.K. and S.S. acknowledge funding from the German Research Foundation (DFG) for the projects BigPlantSens (project no. 444524904) and PANOPS (project no. 504978936). Lastly, M.E.-A. is supported by the Office of the Director of the National Institutes of Health’s Early Investigator Award (1DP5OD029506-01), the U.S. Department of Energy, Office of Biological and Environmental Research (DE-SC0021286), and by the U.S. National Science Foundation's DBI Biology Integration Institute WALII (Water and Life Interface Institute, 2213983). Compute for this project was performed on the Calc cluster at Carnegie Science and the Stanford SC Compute Cluster.

\small
\bibliography{aaai25}

\section{Supplemental Material}
\setcounter{figure}{0}
\renewcommand{\thefigure}{A\arabic{figure}}

\subsection{Building the DivShift-NAWC Dataset}

\begin{figure}
    \centering
    \includegraphics[width=.9\linewidth]{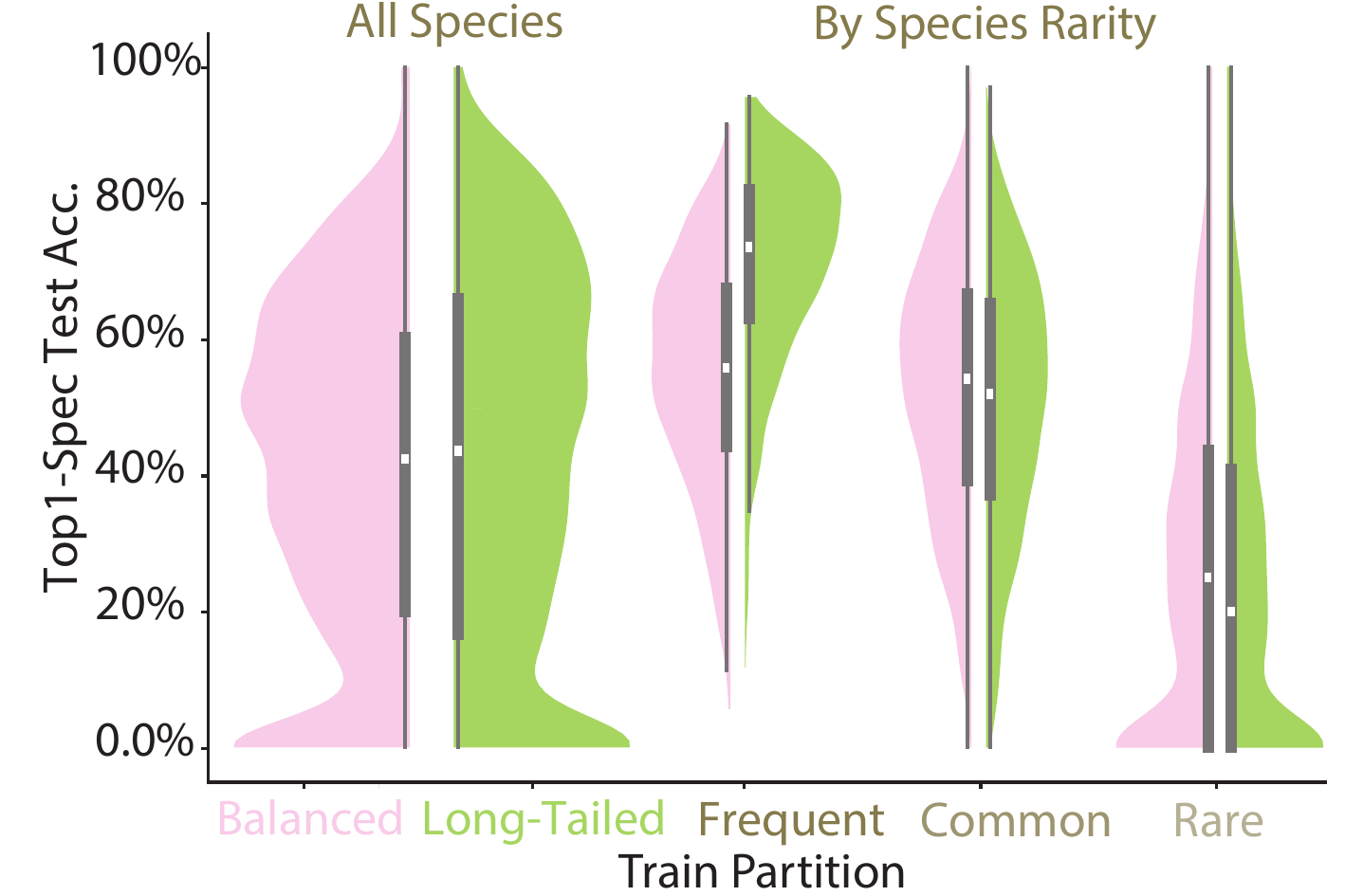}
    \caption{Class-balanced training help in most cases (All Species), but at the cost of common species performance (By Species Rarity).}
    \label{fig:taxanomic}
\end{figure}

Observations were downloaded from the iNaturalist Open Data repository \cite{inats3}. Only research-grade or observations in need of ID were kept. Observations were further filtered to those with a positional accuracy of under 120 m to ensure that spatial associations with geographic variables like climate and habitat type were accurate. Spatial and temporal biases can be taxa-specific \cite{cooper2014there, barve2020methods}, thus given that many plant communities have been undersampled in the past \cite{di2021observing}, we chose to only work with observations of plants, specifically vascular plants (tracheophyta). After filtering to vascular plants, we only kept observations from the years 2019-2023 that fell within the administrative boundaries of the states of Alaska, Yukon, British Columbia, Washington, Oregon, California, Nevada, Arizona, Baja California, Baja California Sur, and Sonora. We further rolled up subspecies, varieties, and phenotypes to the species level to ensure a more uniform intra-class diversity. Lastly, we removed any species not observed in at least two years and only kept species with at least 15 observations. This left us with 3.9 million unique observations, of which 64\% are research grade and labeled with 7,607 unique species (Table \ref{table:Splits}). For each of these observations, we downloaded all available photos per-observation from the iNaturalist Open Data Repository \cite{inats3}, leaving us with 7.3 million unique images of plants.

\begin{figure}
    \centering
    \includegraphics[width=.9\linewidth]{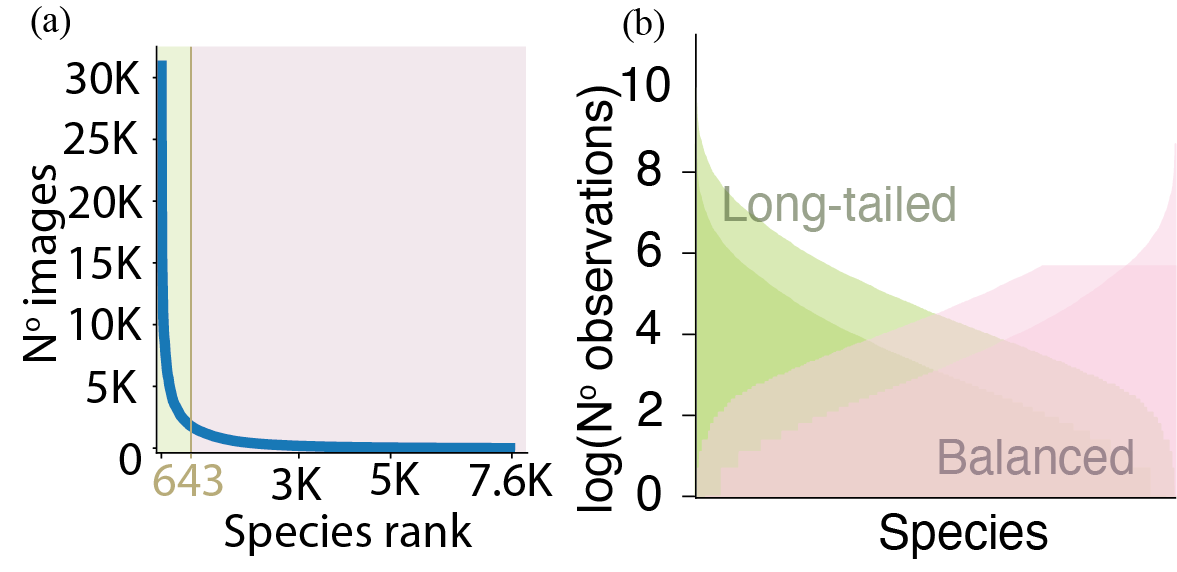}
    \caption{(a) Biodiversity data exhibits a commonness of rarity, where less than 20\% of DivShift-NAWC species possess 80\% of the examples (green) while more than 80\% of species share only less than 20\% of the dataset examples (red). (b) Many machine learning datasets subsample observations from common classes (pink) to create a more balanced dataset than expectation (green).}
    \label{fig:taxanomic-intro}
\end{figure}

iNaturalist data provides crucial and useful information about each image, such as the latitude, longitude, date, and observer~\cite{inat_review}. Using this information, for each observation we added more geologically-relevant data for each image, specifically L2 and L3 ecoregion, 19 current-day WorldClim bioclim variables, land use type, soil type, and Human Footprint data~\cite{ecoregions, WorldClim, landsat_land_use, nachtergaele2023harmonized, HumanFootprint}.

\begin{figure*}
    \centering
    \includegraphics[width=.45\linewidth]{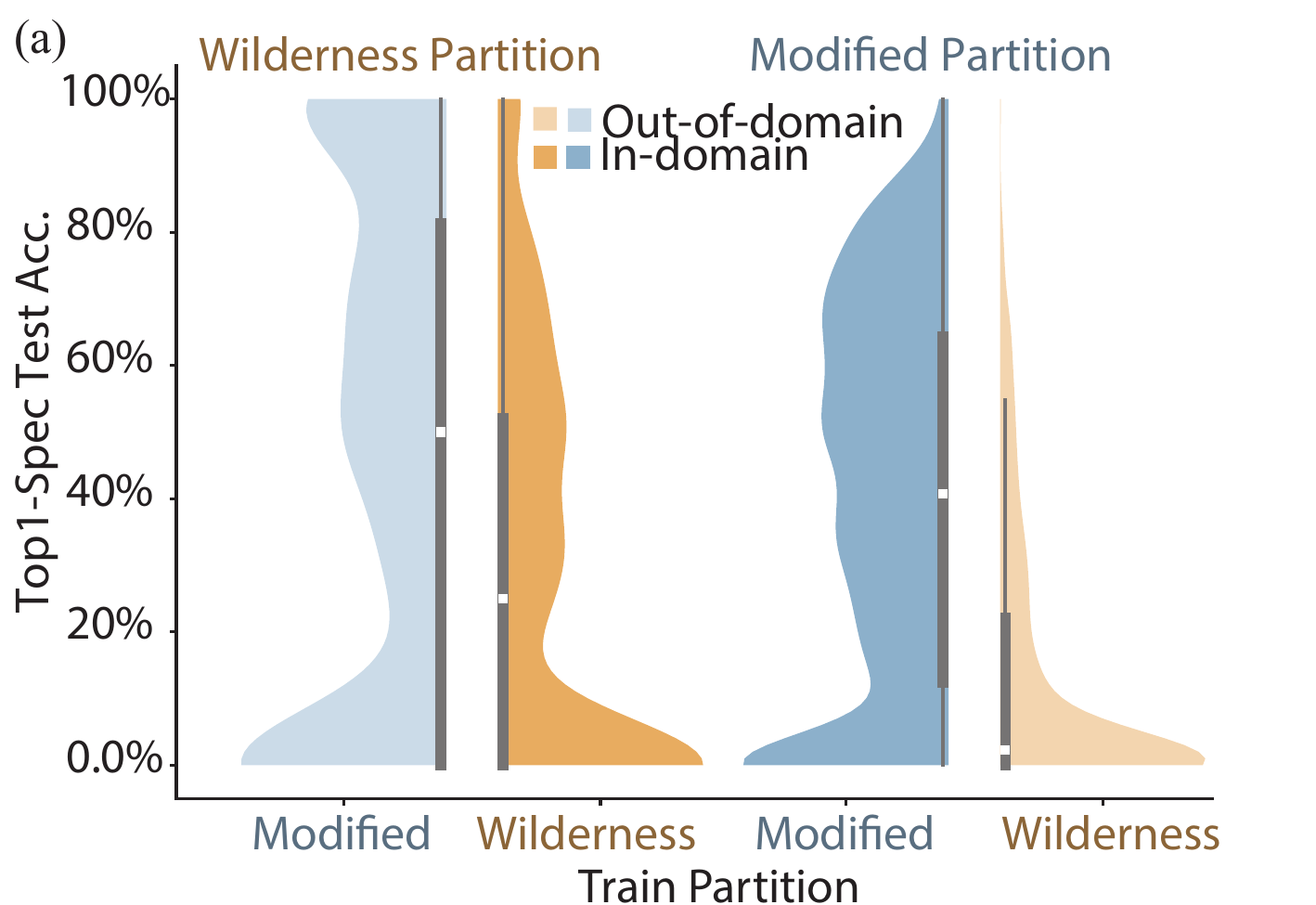}
    \includegraphics[width=.45\linewidth]{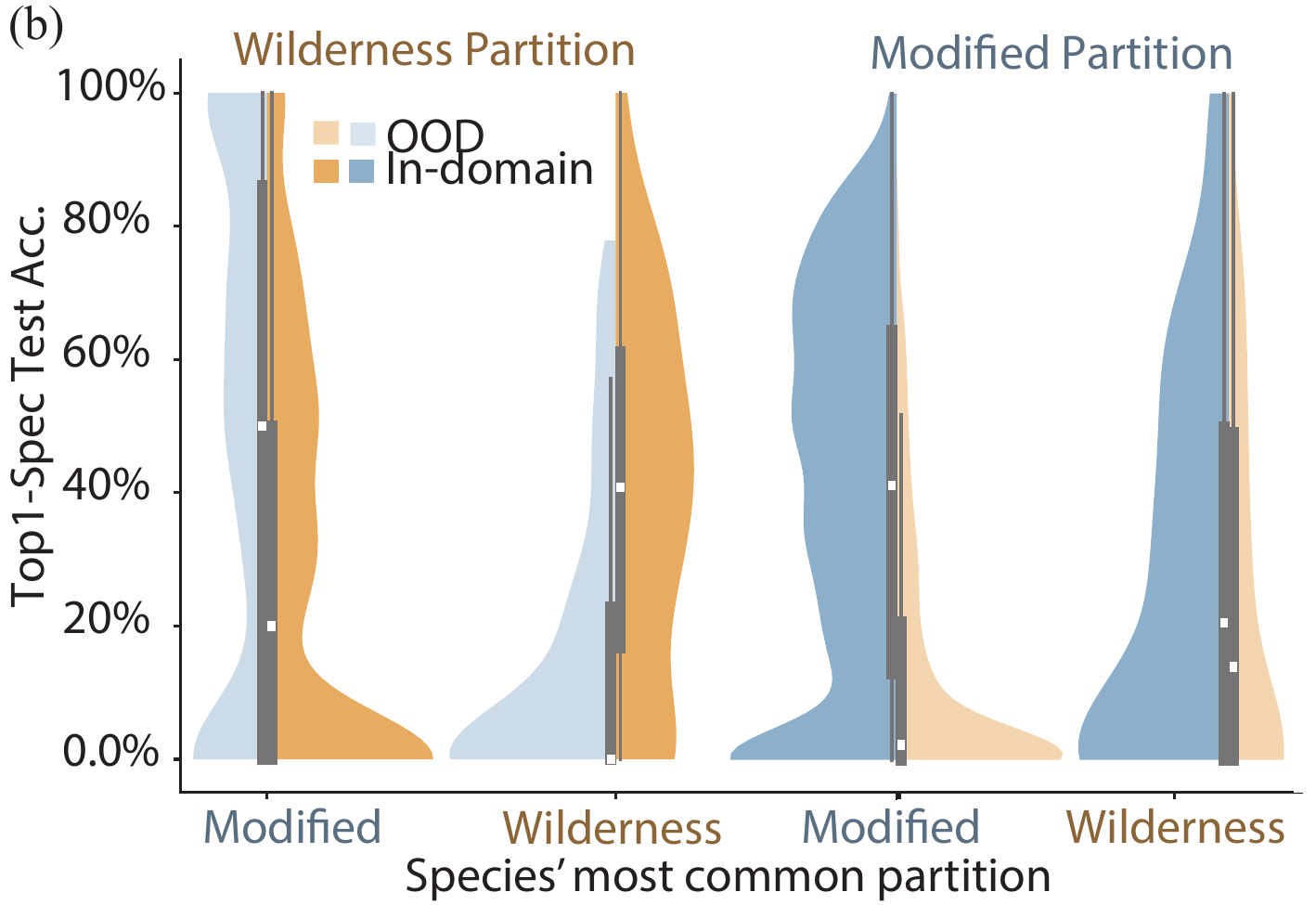}
    \caption{(a) Performance skews to habited areas, but  (b) the wildest species are left behind.}
    \label{fig:spatial}
\end{figure*}

For the purposes of comparing the partition distribution shift to performance shifts in this work, we further filter the DivShift-NAWC dataset to only use research-grade observations. However, the public version of the dataset contains all observation---both research-grade and in need of ID---and will hopefully serve as a platform to additionally develop self-supervised methods these volunteer-collected biodiversity data.

\begin{figure}
    \centering
    \includegraphics[width=.9\linewidth]{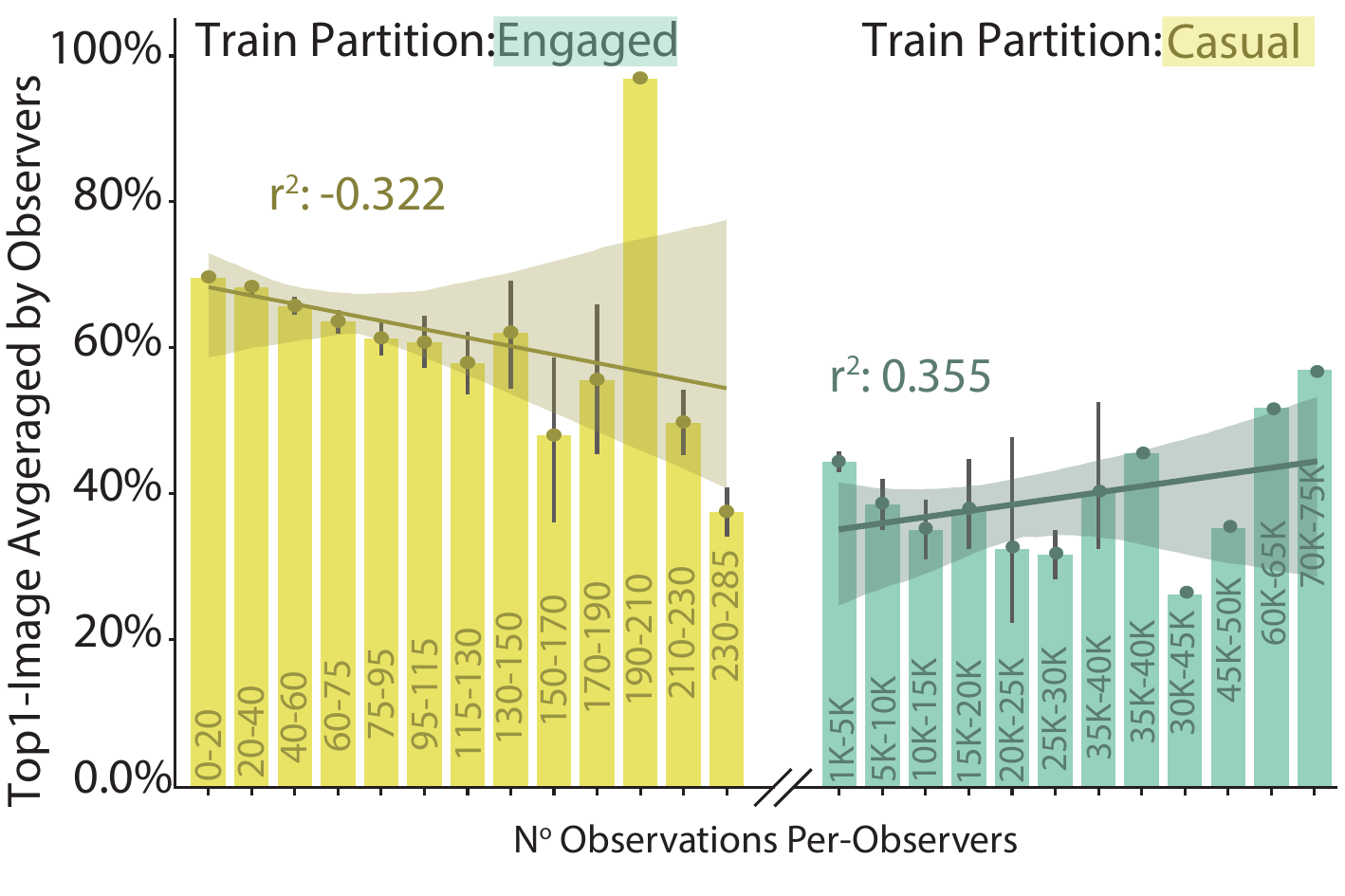}
    \caption{Seasoned observers (Engaged train partition) record better data than inexperienced and occasional users (Casual train partition).}
    \label{fig:quality}
\end{figure}

\subsubsection{Train / Test Split Designation of Partitions}
Each bias partitions was further split randomly into 80\% train and 20\% test using numpy's  implementation of uniform random sampling without replacement. To account for noise in this finite sampling setting, on top of the train / test splits provided in the public version of the dataset, we also include the ability to re-sample a given train / test split with a specific random seed, and use this process to provide aggregated estimates of the JSD between partitions.

\begin{figure*}
    \centering
    \includegraphics[width=.45\linewidth]{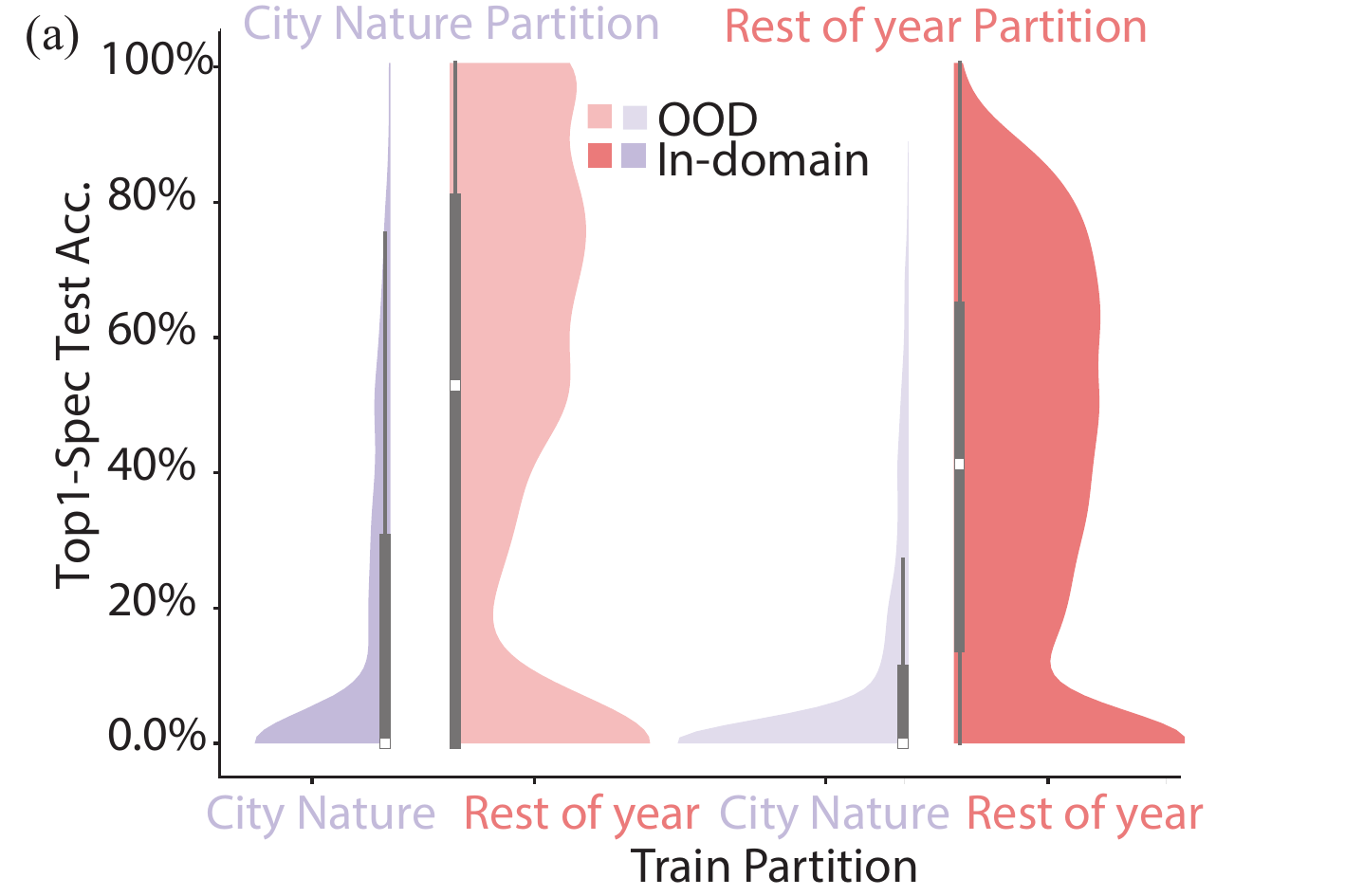}
    \includegraphics[width=.45\linewidth]{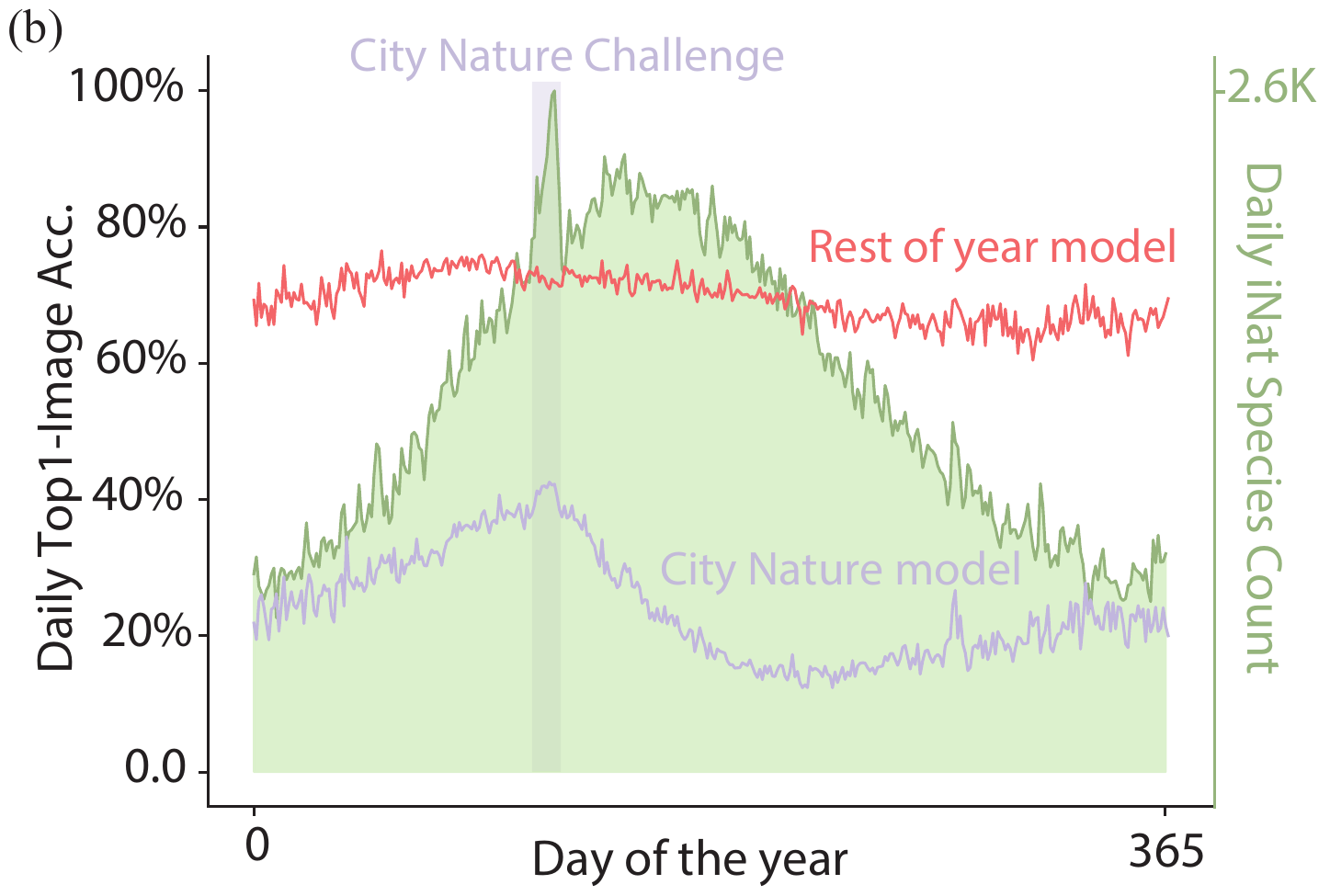}
    \caption{(a) Collection campaigns don't guarantee better data than what's collected across the year, and (b) using City Nature Challenge data alone leads to overfitting.}
    \label{fig:temporal}
\end{figure*}

\subsubsection{Measuring Distribution Shift with Jensen-Shannon Distance}
For each bias partition, we measured the Jensen-Shannon Distance (JSD) between the train set of one partition (e.g. for the spatial bias partition, observations in wilderness areas) to both the train and the test partition of the second partition (e.g. for the spatial bias partition, separately the train and test observations in modified areas). We did this for five different random train/test splits for each partition, giving us mean and error bars reported in Table ~\ref{table:jsd-results}. Of the available statistical distance metrics, we chose to report JSD as it has many desirable properties, namely that is a symmetric metric (e.g. the distance from $P_A$ to $P_B$ is the same as from from $P_B$ to $P_A$) and the metric is bounded from $0$ to $1$ when using a log base of 2, meaning its range can be mapped to the range of differences in accuracies for models trained on these data.
JSD was calculated using scipy's distance module's "jensenshannon" function with a log base of 2. JSD was calculated only for classes present in the training split of a given partition.

\subsubsection{Dataset Licensing and Reuse}
Images and observations available through the iNaturalist Open Data program include data with Creative Commons licenses range from CC-BY-NC, CC-BY-NC-SA, CC-BY-ND, CC0, CC-BY-SA, CC-BY, to CC-BY-NC-ND. These images may be reused for non-commercial purposes and by associativity, the DivShift-NAWC dataset is therefore free and open for research purposes and will be made publicly available along with the associated code to build the dataset and train the models. Individual images can be reproduced with proper attribution given per-image, depending on a photo's given license. License information is provided in the DivShift-NAWC dataset under the column titled "license".



\subsection{Evaluation Metrics}
We report standard computer vision Top-1 accuracy, referred to here as per-Top-1 per-observation accuracy. As the DivShift-NAWC is extremely long-tailed, overweighting the contribution of more common classes to Top-1 per-observation accuracy, we also report Top-1 accuracy averaged per-class (sometimes referred to as macro Top-1 accuracy or average recall), which we refer to here as Top-1 per-species accuracy. This metric considers the Top-1 accuracy per-class independently of classes' frequency. We also report Top-1 per-species accuracy broken down for frequent, common, and rare species \cite{AutoArborist}, defining frequent species as those with 300 or more observations (note that this is the cutoff of observations for the iNat2021 baseline partition), rare species as those with fewer than 50 observations, and common species as those with between 50-300 observations. We also introduce a new rarity-weighted Top-1 accuracy that upweights the relative importance of rarer classes and downweights more common ones within a partition:
$$\frac{1}{\sum_{i=1}^{C}{\frac{1}{Sum(y_i)}}}   \cdot \sum_{i=1}^{C} \frac{Acc(y_i, K)}{Sum(y_i)^2} $$
 where $y$ are the model predictions per-class and per-observation, $Acc(y_i, K)$ is the Top-K accuracy for observations of class $i$, and $Sum(y_i)$ is the number of observations of class $i$ for partition $P_{test}$. 
This metric can be thought of as an inverse of Top-K per-observation accuracy, where rarer classes in a test partition are upweighted and more common classes in a test partition are downweighted. These rarer species may be harder or easier to classify depending on their frequency in the training dataset. Lastly, to modulate the effects of spatial biases, we also calculate per-L2 ecoregion and per-land use category top-1 accuracy \cite{di2021observing}. These accuracies are identical to species Top-K accuracy, except instead of calculating the accuracy per-label class and then averaging across classes, we calculate the Top-K accuracy for all images that fall within a given ecoregion or land-use category, then average those accuracies across the categories.

\subsection{Extended Related Works}
\subsubsection{Definitions of Bias in Biodiversity Data}

The taxonomy, attribution of, and even fundamental definitions of bias in biodiversity data is an active area of study \cite{carlen2024framework, isaac2015bias, isaac2014statistics, di2021observing}. Isaac et. al. defines bias as a property of the observation sampling process, specifically as "variation in recorder activity" \cite{isaac2014statistics}, and acknowledged four forms of bias: non-biological variation in number of observations over time, non-biological variation across space, variation in observation collection effort per-visit, and variation in detectability of organisms \cite{isaac2014statistics, isaac2015bias}. Meanwhile, Di Cecco et. al. partitions biases into spatial, temporal, taxonomic, and user activity level bias \cite{di2021observing}. Lastly, Carlen et. al. defines bias as "an uneven or disproportionate representation of a particular subject or variable within the larger group" \cite{carlen2024framework}, and further categorizes biases that affect observers (referred to as "filters", namely participation, detection, sampling, and preference) and the downstream biases resultant in biodiversity data (such as spatial and temporal) \cite{carlen2024framework}. Carlen et. al. explicitly highlight how sociopolitical biases (referred to as "unconscious bias") strongly affect the participation filter \cite{carlen2024framework}, and importantly Carlen et. al. acknowledges that the there are further intersectional interactions between these biases \cite{carlen2024framework}. For the purposes of this work, we adopt the four definitions of bias from Di Cecco et. al. and additionally include effects of the participation filter from Carlen et. al. as a fifth sociopolitical bias.

\subsubsection{Volunteer-Based Data for Biodiversity Monitoring}
There are a plethora of participatory science platforms and collection strategies for aggregating expert- and volunteer-collected (also known as citizen science and community science) biodiversity datasets. Briefly, these include observation platforms like the Global Biodiversity Information Facility \cite{gbif} which allow researchers and registered members of the public to upload geolocated and timestamped observations for both individual species observations and community checklists;  plant-specific \cite{calflora} and bird-specific \cite{christmasbird} databases; easy-to-use apps targeted for casual users like iNaturalist \cite{inat}, Pl@ntNet \cite{pantnet}, and eBird \cite{sullivan2009ebird} which allow users to upload geolocated and timestamped photos of individuals or checklists in real-time, identify them, and share them to publicly; and more structured and specialized checklists like relevés \cite{cnps}, targeted collection campaigns focused on specific taxa \cite{geldmann2016}, and eDNA soil collection campaigns \cite{lin2021landscape}. 

Community engagement projects built around these strategies have in turn enabled a wide array of novel and impactful biodiversity monitoring breakthroughs, such as species and habitat monitoring \cite{cornwell2012co, fontaine2022scientific, lehtiniemi2020citizen, hawthorne2015mapping, callaghan2019using}, tracking invasive species spread \cite{werenkraut2020citizen, moulin2020citizen, hiller2019case, gallo2011creating}, detecting new populations of species \cite{wilson2020more, smith2022citizen}, rediscovering cryptic species \cite{wesener2018integrative}, quantifying and monitoring species richness \cite{neyens2019mapping, callaghan_capitalizing_2020}, quantifying anthropogenic biodiversity changes \cite{leong2019citizen, forister2021fewer, girish2022community, rapacciuolo2021deriving, champion2018rapid}, understanding species interactions \cite{gazdic2019inaturalist, lopez2020insights, marin-gomez_assessing_2022}, characterizing within-species diversity and behavior \cite{drury2019continent, barve2020methods}, estimating species' population sizes \cite{ver2021species, horns2018using, walker2017using, van2013opportunistic}, tracking ecological disaster recovery efforts \cite{mccormick2012after}, and aiding conservation decisions \cite{sullivan2017using, robinson2018using, loss2015linkingg}. 

These projects are now considered to be an essential tool for reaching conservation goals across the world \cite{brown2019potential, devictor2010beyond, chandler2017contribution, aceves2015citizen, theobald2015global, pocock2018vision, mckinley2017citizen}. From these data collections and sampling strategies, we focus specifically on iNaturalist as it is the largest data collection with linked images for almost every observation (excepting some bird observations identified by an audio recording of their call) \cite{inat_review}.

\subsubsection{Spatial Bias in Volunteer-Collected Biodiversity Data}
Additional drivers of spatial bias in volunteer-collected biodiversity data include participants sampling closer to home \cite{gratzer2021and, mcgoff2017finding} and differential preferences for protected wilderness versus urban greenspaces \cite{backstrom2024estimating, la2024data, di2021observing, dimson_who_2023}. These spatial biases can affect inferences about 
demographic changes \cite{boakes2010distorted, backstrom2024estimating}, biodiversity changes \cite{rapacciuolo2021deriving}, and the utility of these data for conservation planning \cite{botts2011geographic}. Various methods have been proposed and tested to mitigate the effects of spatial bias \cite{zizka2021sampbias, jacobs2017completeness}, mainly for species distribution modeling \cite{van2021impact, steen2021spatial, steen2019evaluation, tang2021modeling, johnston2020estimating}. 

\subsubsection{Temporal Bias in Volunteer-Collected Biodiversity Data}
Additional drivers of temporal bias in volunteer-collected biodiversity data include the year-over-year rise in popularity of participatory science platforms \cite{di2021observing, backstrom2024estimating, dimson_who_2023}, relative ease of observing on weekends versus the workweek \cite{di2021observing,courter2013weekend, cooper2014there}, and the COVID-19 pandemic \cite{sweet2022covid, sanchez2021differential, crimmins_covid-19_2021}. These temporal biases can make it difficult to accurately assess bird migration patterns \cite{steiner2022changes}, changes in species distributions  \cite{bowler2022temporal, daniel2023temporal}, population declines \cite{kamp2016unstructured}, and flowering time \cite{park2021scale, belitz2020accuracy} from these data. Methods do exist to mitigate these effects \cite{boyd2022robitt}, but mainly for estimating demographic changes over time \cite{backstrom2024estimating, fink2023double}. 

\begin{figure}
    \centering
    \includegraphics[width=.9\linewidth]{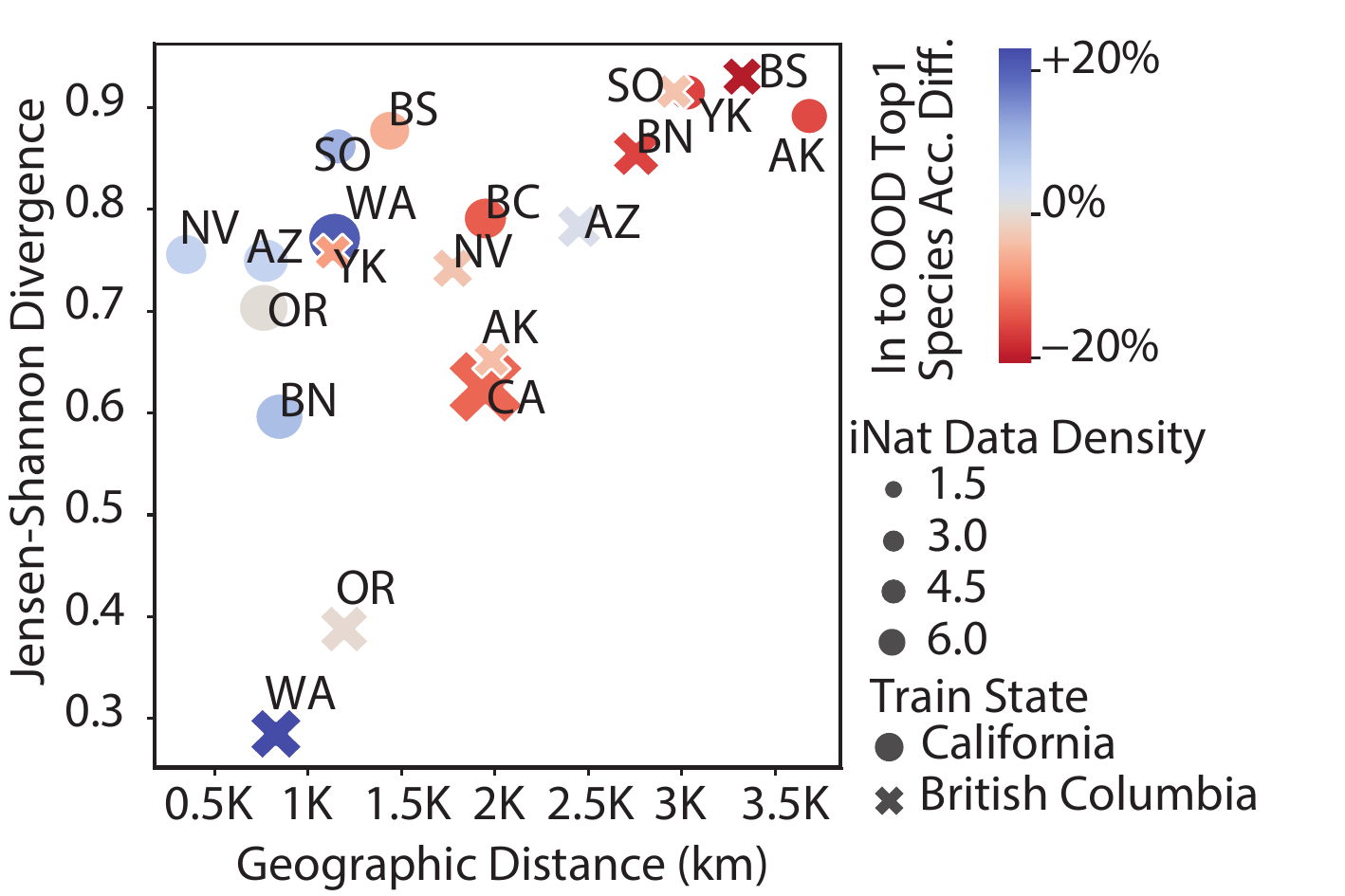}
    \caption{Geographical and ecological distance correlate along with accuracy.}
    \label{fig:socio-political}
\end{figure}

\subsubsection{Taxonomic Bias in Volunteer-Collected Biodiversity Data}
Additional drivers of taxonomic biases in volunteer-collected datasets stem in part from difficulty detecting some species \cite{aristeidou2021exploring, unger2021inaturalist}, over-representation of species more common to urban environments or roadsides \cite{ward2014understanding, mair2016explaining}, and the amount of technical specialization needed to identify species \cite{mcmullin2022assessment, hochmair2020evaluating, boakes_patterns_2016}, especially threatened ones \cite{deacon2023overcoming}. There are additional marked differences between the location where certain taxa are observed, most notably with birding hotspots more often in preserved areas \cite{la2024data} while iNaturalist data more broadly tends to be observed in disturbed environments \cite{di2021observing, dimson_who_2023}. These differences make it difficult to generalize conclusions from one taxa or one platform to another, and mitigation solutions generally tend to focus on the design of participatory science projects \cite{deacon2023overcoming} as opposed to explicit modeling.

\subsubsection{Observation Quality Bias in Volunteer-Collected Biodiversity Data}
Observation quality---defined here as how representative a collection of observations are of the underlying biodiversity of an area---are driven in part by who is observing. For bird surveys, a small but highly-specialized subset of observers contribute the most observations \cite{rosenblatt2022highly}, and more generally more active users tend to observe more species in more diverse habitats \cite{di2021observing}. Observer behavior and observation quality also differ based on whether observers are local residents or visitors \cite{dimson_who_2023}. Filtering observations from the most active users or their most active days \cite{van2021impact, milanesi2020observer, boakes_patterns_2016} is the main approach for mitigating these effects currently.

\subsubsection{Sociopolitical Bias in Volunteer-Collected Biodiversity Data} Lastly, sociopolitical factors influence who observes where. This induces a skew towards whiter, wealthier, older, and more educated observers \cite{mac2020citizens,pateman2021diversity} and fewer observations in areas and communities of environmental justice concern \cite{blake2020demographics}, fewer observers in historically redlined districts or communities of color in the U.S. \cite{ellis2023historical, mahmoudi2022mapping}, fewer observations in lower GDP countries \cite{deacon2023overcoming},
differential access to green spaces \cite{chen2022contrasting}, and conservation and land management policy differences across political boundaries \cite{dallimer2015socio}. Potential solutions include more broad structural reform of participatory science platforms and projects \cite{carlen2024framework, burgess2017science, cooper2023equitable,soleri2016finding, pandya2012framework}, but little work exists to account for these differences in a modeling context.

\subsubsection{Computer Vision Approaches to Domain Shift}
From the computer vision community, a variety of datasets that test model performance under domain shift exist \cite{koh2021wilds, Wild-Time, sagawa2021extending}. Modeling strategies to minimize domain shift effects include distributionally robust optimization \cite{sagawa2019distributionally}, modifying augmentations between pre-training and fine-tuning \cite{qu2024connect}, and using attentional-biased stochastic gradient descent \cite{qi2022attentional}. For long-tailed data like commonly found with biodiversity data collections, strategies include k-positive contrastive learning for long-tailed settings \cite{kang2020exploring}, minority class oversampling \cite{park2022majority}, and hard conditional negative sampling \cite{wu2020conditional}. 

\end{document}